\documentclass[runningheads]{llncs}

\usepackage{graphicx}
\usepackage{comment}
\usepackage{amssymb} % define this before the line numbering.
\usepackage{color}
\usepackage{amsmath}
\usepackage{mathtools}

\usepackage[pagebackref,breaklinks,colorlinks]{hyperref}
\usepackage[capitalize]{cleveref}

\usepackage{graphicx}
\usepackage{amsmath}
\usepackage{amssymb}
\usepackage{booktabs}
\usepackage{bm}
\usepackage{multirow}
\usepackage[table,xcdraw]{xcolor}

\usepackage{overpic}
\usepackage{enumitem} %< control spacing in itemize/enumerate/...
\usepackage{overpic} %< add raw math symbols to figures
\usepackage{color}
\definecolor{turquoise}{cmyk}{0.65,0,0.1,0.3}
\definecolor{purple}{rgb}{0.65,0,0.65}
\definecolor{dark_green}{rgb}{0, 0.5, 0}
\definecolor{orange}{rgb}{0.8, 0.6, 0.2}
\definecolor{red}{rgb}{0.8, 0.2, 0.2}
\definecolor{darkred}{rgb}{0.6, 0.1, 0.05}
\definecolor{blueish}{rgb}{0.0, 0.3, .6}
\definecolor{light_gray}{rgb}{0.7, 0.7, .7}
\definecolor{pink}{rgb}{1, 0, 1}
\definecolor{greyblue}{rgb}{0.25, 0.25, 1}
\definecolor{greenP}{RGB}{118, 170, 74}
\definecolor{blueQ}{RGB}{129, 159, 228}
\definecolor{orangeD}{RGB}{227, 187, 81}
\definecolor{grayF}{RGB}{116, 116, 116}

\newif\ifshowcomments
\showcommentstrue % uncomment to show the comments

\ifshowcomments
    \newcommand{\todo}[1]{{\color{red}#1}}
    \newcommand{\TODO}[1]{\textbf{\color{red}[TODO: #1]}}
    
    \newcommand{\ea}[1]{{\color{blueish}#1}}
    \newcommand{\EA}[1]{{\color{blueish}{\bf [EA: #1]}}}
    \newcommand{\Ea}[1]{\marginpar{\tiny{\textcolor{blueish}{#1}}}}
    \newcommand{\AS}[1]{{\color{red} \bf [AS: #1]}}
    \newcommand{\MK}[1]{{\color{purple} \bf [MK: #1]}}
    \newcommand{\oh}[1]{{\color{orange}#1}}
    \newcommand{\Oh}[1]{\marginpar{\tiny{\textcolor{orange}{#1}}}}
    \newcommand{\OH}[1]{{\color{orange}{\bf [OH: #1]}}}
    \newcommand{\sm}[1]{{\color{dark_green}#1}}
    \newcommand{\Sm}[1]{\marginpar{\tiny{\textcolor{dark_green}{#1}}}}
    \newcommand{\SM}[1]{{\color{dark_green}{\bf [SM: #1]}}}
\else
    \newcommand{\todo}[1]{\unskip}
    \newcommand{\TODO}[1]{\unskip}
    \newcommand{\ea}[1]{\unskip}
    \newcommand{\EA}[1]{\unskip}
    \newcommand{\Ea}[1]{\unskip}
    \newcommand{\AS}[1]{\unskip}
    \newcommand{\MK}[1]{\unskip}
    \newcommand{\oh}[1]{\unskip}
    \newcommand{\Oh}[1]{\unskip}
    \newcommand{\OH}[1]{\unskip}
    \newcommand{\sm}[1]{\unskip}
    \newcommand{\Sm}[1]{\unskip}
    \newcommand{\SM}[1]{\unskip}
\fi

\DeclareMathOperator*{\argmin}{arg\,min}

\newcommand{\real}{\mathbb{R}}
\newcommand{\gaussian}{\mathcal{N}}
\newcommand{\gaussianpri}{\mathcal{N}(\vecb{0}, \vecb{I})}
\newcommand{\vecb}[1]{\textbf{#1}}

\newcommand{\vecQ}[1]{\textbf{#1}^{(q)}}
\newcommand{\vecP}[1]{\textbf{#1}^{(p)}}
\newcommand{\bmQ}[1]{\bm{#1}^{(q)}}
\newcommand{\bmP}[1]{\bm{#1}^{(p)}}
\newcommand{\norm}[1]{\left\lVert#1\right\rVert}

\newcommand{\Fig}[1]{Fig.~\ref{fig:#1}}

\newcommand{\Tab}[1]{Tab.~\ref{tab:#1}}

\newcommand{\Eq}[1]{Eq.~\ref{eq:#1}}

\newcommand{\Sec}[1]{Sec.~\ref{sec:#1}}

\newcommand{\Supp}{supp. mat.}

\newcommand{\model}{LiP-Flow}
\newcommand{\modelname}{LiP-Flow~}
\newcommand{\modelFlow}{LiP-Flow~}
\newcommand{\modelKL}{LiP-KL~}
\newcommand{\modelKLnospace}{LiP-KL}

\usepackage{blindtext}

\renewcommand{\paragraph}[1]{\vspace{1em}\noindent\textbf{#1}.}

\newcommand{\ie}{\textit{i.e.}}
\newcommand{\eg}{\textit{e.g.}}
\newcommand{\etal}{\textit{et al.}}
\usepackage{tikz}
\usepackage{wrapfig}
\usepackage{sidecap}
\usepackage{nicematrix}

\usepackage[accsupp]{axessibility}  % Improves PDF readability for those with disabilities.

\renewcommand{\textfloatsep}{0.5ex}

\makeatletter
\def\@fnsymbol#1{\ensuremath{\ifcase#1\or \mathsection\or \dagger\or \ddagger\or
   \mathparagraph\or \|\or **\or \dagger\dagger
   \or \ddagger\ddagger \else\@ctrerr\fi}}
    \makeatother

\begin{document}

\pagestyle{headings}
\mainmatter

\title{\large{LiP-Flow: Learning Inference-time Priors for Codec Avatars via Normalizing Flows in Latent Space}}

\author{Emre Aksan\thanks{This work was performed during an internship at Meta Reality Labs Research.}\inst{1}
\and Shugao Ma\inst{3}
\and Akin Caliskan\textsuperscript{$\mathsection$}\inst{2}
\and Stanislav Pidhorskyi\inst{3}
\and \\ Alexander Richard\inst{3}
\and Shih-En Wei\inst{3} 
\and Jason Saragih\inst{3} 
\and Otmar Hilliges\inst{1}
}
\institute{ETH Zürich, Department of Computer Science 
\and CVSSP, University of Surrey 
\and Meta Reality Labs Research}

\authorrunning{E. Aksan et al.}
\titlerunning{LiP-Flow: Learning Inference-time Priors via Normalizing Flows}

\maketitle
\begin{abstract}
Neural face avatars that are trained from multi-view data captured in camera domes can produce photo-realistic 3D reconstructions. However, at inference time, they must be driven by limited inputs such as partial views recorded by headset-mounted cameras or a front-facing camera, and sparse facial landmarks. To mitigate this asymmetry, we introduce a prior model that is conditioned on the runtime inputs and tie this prior space to the 3D face model via a normalizing flow in the latent space. Our proposed model, LiP-Flow, consists of two encoders that learn representations from the rich training-time and impoverished inference-time observations. A normalizing flow bridges the two representation spaces and transforms latent samples from one domain to another, allowing us to define a latent likelihood objective. We trained our model end-to-end to maximize the similarity of both representation spaces and the reconstruction quality, making the 3D face model aware of the limited driving signals. We conduct extensive evaluations where the latent codes are optimized to reconstruct 3D avatars from partial or sparse observations. We show that our approach leads to an expressive and effective prior, capturing facial dynamics and subtle expressions better. Check out our \href{https://youtu.be/uvZ4npQV5GU}{video} for an overview.
\end{abstract}

\section{Introduction}
\label{sec:intro}

\begin{figure}[t]
  \begin{minipage}[c]{0.55\textwidth}
    \includegraphics[width=\textwidth, trim={20 250 530 30},clip]{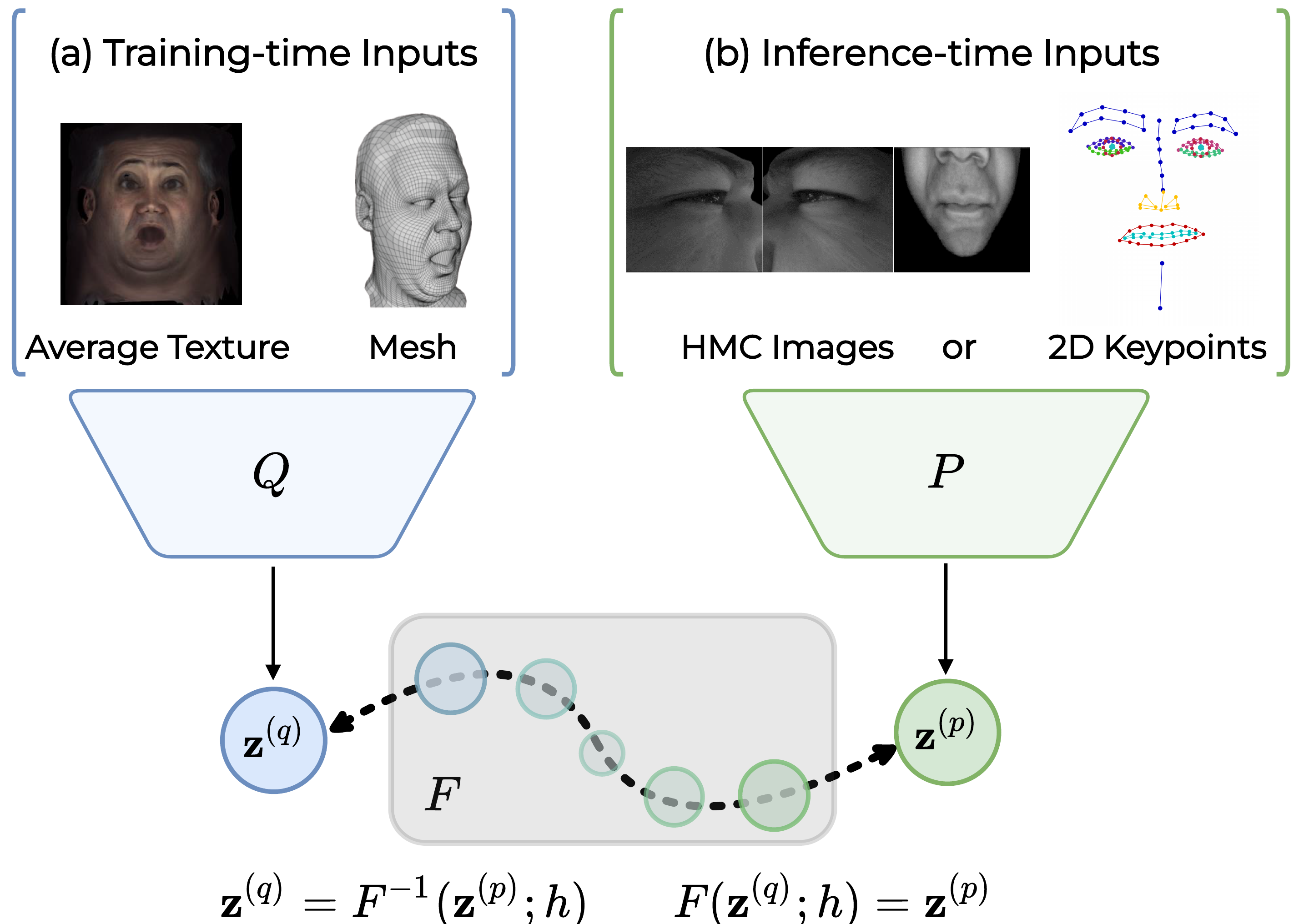}
  \end{minipage} \hfill
  \begin{minipage}[c]{0.43\textwidth}
    \vspace{-10pt}
    \caption{
        \textbf{Information asymmetry} between the (a) training- and (b) inference-time observations. We learn a \textcolor{grayF}{normalizing flow model ($F$)} across latent spaces, minimizing the discrepancy between the 
        base network's Deep Appearance Model (DAM) encoder \cite{lombardi2018deep}, \textcolor{blueQ}{DAM-encoder (Q)}, and an \textcolor{greenP}{inference-time encoder ($P$)}, namely the HMC-encoder or KPT-encoder conditioned on partial head-mounted camera (HMC) images or sparse 2D keypoints (KPT), respectively. 
    }
    \label{fig:teaser}
  \end{minipage}
  % \vspace{0.2cm}
\end{figure}
\vspace{-0.2cm} 
VR telepresence promises immersive social interactions. To experience the presence of others as genuine, such a system must provide photo-realistic 3D renderings, capture fine-grained details such as pores and hair, and produce subtle facial dynamics. 
Codec Avatars \cite{lombardi2018deep,wei2019vr} have been shown to be a promising direction towards this goal. Such systems implement 3D avatars via the decoder network of a variational auto-encoder (VAE) \cite{kingma2013auto}. Per subject, high-quality 3D face avatars are trained with multi-view imagery from a 40+ camera dome \cite{lombardi2018deep}. 

In prior work \cite{bi2021deep,chu2020expressive,lombardi2018deep,ma2021pixel,schwartz2020eyes}, the network learns to decode latent codes into shape and view-conditioned appearance. Importantly, each latent code is a projection of the geometry and unwrapped texture which are only available during \emph{training} (\Fig{teaser}-a). At \emph{inference-time}, the driving signal is impoverished -- consisting of a set of partial face images captured from cameras mounted on a VR headset \cite{wei2019vr}, facial landmarks \cite{chandran2020semantic} (\Fig{teaser}-b) or a single-view from a frontal camera \cite{cao2021real}.
To maximize the reconstruction quality, the avatar model is often trained in isolation, and typically a separate run-time encoder for the driving signal is used.. 
We show that this separate training scheme limits performance since the constrained run-time conditions 
are not considered during decoder training.

\begin{figure*}[t]
\begin{center}
\includegraphics[width=0.9\linewidth, trim={30 950 150 0},clip]{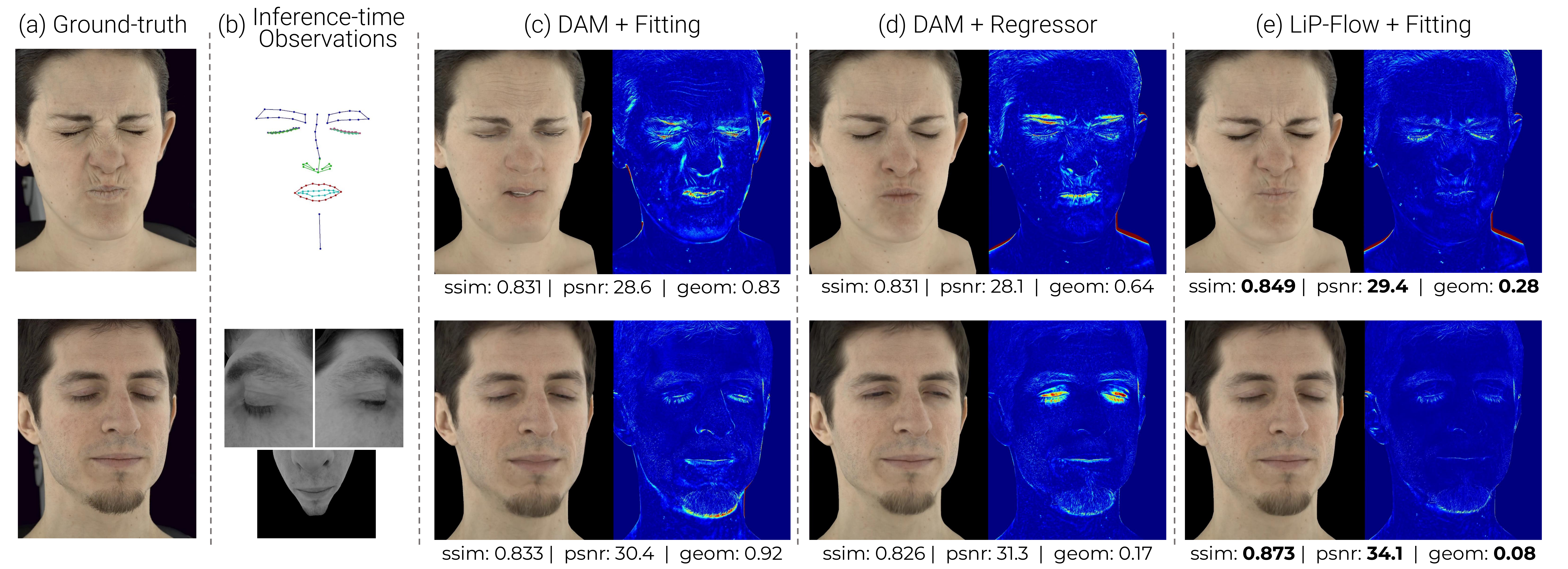}
\end{center}
\vspace{-0.5cm}
\caption{
\textbf{Driving a 3D avatar with impoverished inputs}. 
% We provide error maps between the ground-truth (a) and the reconstructions, and report image and geometry errors. 
For the DAM (c) and our proposed model \modelname(e), we first optimize latent codes to reconstruct the inference-time observations (b), and then reconstruct the full face. % sparse facial landmarks or partial HMC images (b). 
In (d), we train a regressor to directly predict latent codes from the inference-time inputs (b) by using the pretrained DAM-encoder ($Q$) in (c). 
% (Top) For the peak expressions, the difference between our \modelname and the baselines is visible in the reconstructed images. (Bottom) Even for the less challenging neutral expressions, the difference is notable in the error maps and the loss metrics, though less so in the image.
}
\label{fig:iterative_fitting}
% \vspace{0.1cm}
\end{figure*}

In this work, we introduce \model, a learned inference-time prior, based on normalizing flows, to bridge this information gap for 3D face models.
Specifically, we incorporate a prior model into the Deep Appearance Model (DAM) pipeline \cite{lombardi2018deep} with minimal modifications such that the decoder's high reconstruction quality is preserved.
We make use of this prior model in \emph{inference-time} optimization tasks.
This allows us to optimize latent codes to capture facial dynamics and subtle expressions even when only impoverished inputs are available (\Fig{iterative_fitting}). 
It is effective even on less challenging neutral expressions, while it is more pronounced on peak expressions.
We simulate the run-time conditions via sparse 2D keypoints (KPT) and synthetically generated partial head-mounted camera (HMC) images, allowing to define (virtual) correspondences between training and inference data and to run tightly controlled evaluations in all settings.

Data-driven priors have many uses in under-constrained problems. 
Given partial or imperfect observations, the task is the reconstruction of a clean signal or inference of the latent factors leading to it. 
This task has been studied in the context of related areas, amongst others 3D lifting \cite{bogo2016keep,pavlakos2019expressive}, 3D pose and shape estimation from images \cite{chen2020category,rempe2021humor,spurr2018cross,zhou20153d}, and refining \cite{li2021task}, image editing \cite{bau2020semantic,pan2021exploiting}, or prediction of the next step from the past in sequence models \cite{aksan2019stcn,aksan2018deepwriting,chu2020expressive,lai2018stochastic,rempe2021humor}. Our focus lies particularly on learning latent priors conditioned on limited observations \cite{aksan2019stcn,chung2015recurrent,rempe2021humor} and using prior models for inference in a more compact representation space \cite{chen2020category,kolotouros2021probabilistic,pavlakos2019expressive,rempe2021humor}. In our work, we learn a prior that is conditioned on the HMC views or facial landmarks. We aim to regularize the main avatar model by incorporating the run-time observations into the training pipeline and increase reconstruction quality in inference-time optimization.

Our work introduces a novel approach for learning of conditional priors. 
To minimize the discrepancy between the training- and inference-time representations, we learn a mapping across the two latent spaces via a conditional normalizing flow (\Fig{teaser}) \cite{winkler2019learning,bhattacharyya2019conditional}. 
It is noteworthy to mention that both the training- and inference-time encoders, our latent flow network as well as the decoder are jointly trained.  
Our training objective consists of reconstruction terms for appearance and geometry, and a latent likelihood term, replacing the KL-divergence (KL-D) term in the VAE framework. Specifically, we optimize the likelihood of the latent samples from the DAM-encoder ($Q$) under the learned inference-time prior (LiP) distribution estimated from the HMC views or 2D keypoints. This is enabled by the bijective nature of the underlying flow, allowing us to map samples from one space to another.
We note that the information asymmetry is pronounced in our task both in terms of modality and amount of carried information (\ie, geometry and texture vs. HMC images or keypoints).
The KL-D objective enforces distribution similarity between the latent-spaces which we show experimentally may be a too strong of an assumption, given the differences in input distributions. 
We relax this assumption by leveraging a normalizing flow to bridge between the latent spaces without enforcing distributional similarity. We show that our approach constitutes a powerful means to deal with such large discrepancies.

We first demonstrate how our latent space formulation improves the reconstruction performance of the base DAM network. Next, we compare different approaches for learning a prior model conditioned on the HMC images or 2D keypoints. We evaluate the quality of the prior models via an analysis-by-synthesis technique \cite{chen2020category,nair2008analysis}. More specifically, we aim to recover the underlying latent code from HMC images or facial landmarks. We treat the pre-trained decoder (\ie, the 3D face model) as a black box renderer to produce an image for a given latent code, view direction, and camera pose. 
In this task, in addition to a reconstruction objective, we also leverage the learned prior via a likelihood objective on the latent codes. 
We show that our flow-based prior model yields latent codes that reconstruct the 3D face faithfully (\Fig{iterative_fitting}). Our contributions can be summarized as follows: (1) a novel approach for learning conditional priors that can be integrated seamlessly into a base architecture for the task of 3D face avatar generation and animation, (2) a more expressive and flexible representation space compared to KL-D driven latent spaces, (3) better reconstruction quality than the baseline approaches in tightly controlled extensive evaluations.

% \clearpage
% \vspace{-0.2cm}
\section{Related work}
\label{sec:related}
\vspace{-0.5cm}
\paragraph{3D Face Appearance Models}
Modeling of 3D human faces for animation has long been an active area of research. Early work models 3D human faces with linear combinations of blendshape vectors representing meshes and texture (\eg \cite{Blanz99}). We refer to Lewis \etal~\cite{Lewis2014PracticeAT} for a review. 
Such models are limited in their expressiveness: a large number of blendshapes are needed to create high-fidelity animations. 
Recent works \cite{buhler2021varitex,feng2021learning} leverages 3D parametric face models to predict texture and shape for novel identities.
Many deep appearance face models utilize deep generative neural networks such as VAEs \cite{Bagautdinov_2018_CVPR,chandran2020semantic,lombardi2018deep,ma2021pixel,tewari17MoFA} or GANs \cite{Abrevaya_2019_ICCV,cheng2019meshgan,Shamai2019,Slossberg_2018_ECCV_Workshops} with facial expressions represented as compact {\em latent codes}. 

In particular {\em Codec Avatars} \cite{lombardi2018deep,ma2021pixel} trained with high resolution multi-view images can photo-realistically render human faces. 
Face animations can also be generated when conditioned on HMC images \cite{chu2020expressive,wei2019vr}, frontal-view images \cite{cao2021real} or even audio \cite{richard2021audio}. However, the significant discrepancy between training- and inference-time inputs is a key challenge for achieving high fidelity facial animation for telepresence applications. To alleviate this problem, we explore novel approaches augmenting the Codec Avatar pipeline via conditional priors..

\vspace{-0.15cm}
\paragraph{Latent Code Optimization} 
Generative models like GANs \cite{goodfellow2014generative} and VAEs \cite{kingma2013auto} can represent the underlying data manifold in a latent space. They have been used as data-driven priors in solving inverse problems, via optimization in the more abstract latent space. 
The standard Gaussian latent prior of a VAE \cite{pavlakos2019expressive} and a hierarchical VAE \cite{li2021task} replace a GMM prior in the output space \cite{bogo2016keep} for modeling and reconstructing 3D human pose. Chen \etal \cite{chen2020category} also use a VAE decoder and regularize the latent codes explicitly by evaluating the latent likelihood under the VAE prior. 
In \cite{bau2020semantic,pan2021exploiting}, GANs and in \cite{guo2019agem}, denoising autoencoders are treated as a block-box sample generator for image in-painting and denoising tasks. Similarly, GANFit \cite{gecer2019ganfit} models facial texture via a GAN, and jointly optimizes the corresponding latent code and parameters of a 3D morphable model. Our work differs from those in that we learn a conditional prior model allowing us to calculate the likelihood of the latent code  during optimization. 
HuMoR \cite{rempe2021humor} also provides a learned prior that affords likelihood evaluation. It is conditioned on the 3D pose in the previous frame and regularized via KL-D. 
Instead, we propose a flow-based approach for learning conditional priors in settings where the inference-time data differs significantly from the training data.

Normalizing flow priors have been proposed for inverse problems \cite{asim2020invertible,whang2021composing,zanfir2020weakly}. In \cite{asim2020invertible}, an invertible Glow model \cite{kingma2018glow} replaces GANs for image inpainting. Similarly in \cite{kolotouros2021probabilistic}, a normalizing flow model is used for 2D-3D lifting. Zanfir \etal \cite{zanfir2020weakly} introduce a 3D human pose prior by learning a mapping between a Gaussian latent space and 3D human poses via normalizing flows. These works make use of the generative nature of unconditional flows and replace GANs or VAEs 1-to-1. 
The LSGM \cite{vahdat2021score} learns an invertible mapping between a standard Normal prior and an encoder space via a diffusion process. In our work, we introduce a different setting where we jointly learn separate representation spaces with a large discrepancy in the inputs and the mapping between them.

% \clearpage

\begin{figure*}[t]
\begin{center}
\includegraphics[width=\linewidth, trim={5 110 10 20},clip]{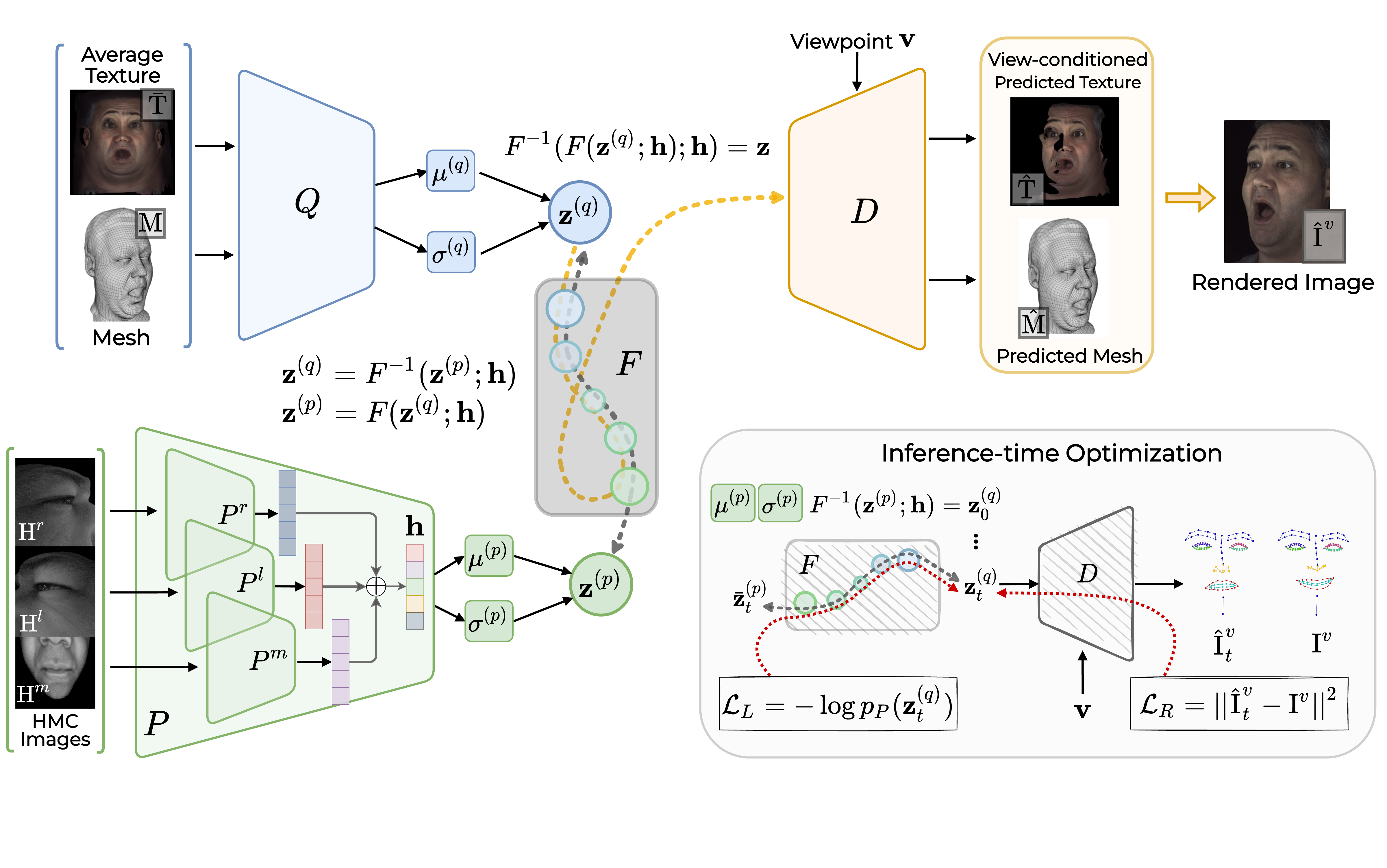}
\end{center}
\vspace{-0.3cm}
\caption{
\textbf{\model~with HMC inputs}. 
Without losing generality, we introduce an \textcolor{greenP}{HMC-encoder ($P$)}, treated as a prior model conditioned on the HMC images, which can be replaced with a \textcolor{greenP}{KPT-encoder ($P$)} conditioned on 2D keypoints. 
The prior model \textcolor{greenP}{($P$)} is trained end-to-end with the base \textcolor{blueQ}{DAM-encoder ($Q$)} and the \textcolor{orangeD}{decoder ($D$)}. The \textcolor{grayF}{bijective function ($F$)} learns to map samples from one space to another. We train the decoder ($D$) with latent samples $\vecQ{z}$ only. To get the reconstruction training signal in the \textcolor{grayF}{flow ($F$)}, the input code follows the flow cycle $\vecb{z} = F^{-1}(F(\vecQ{z}; \vecb{h}); \vecb{h})$. \textbf{Inference-time Optimization}. 
To faithfully reconstruct the 3D face only from runtime inputs, we optimize the latent codes to fit the observations such as HMC images or 2D keypoints.
We first estimate a prior distribution \textcolor{greenP}{($P$)} with the inference-time encoders \textcolor{greenP}{HMC-encoder} or \textcolor{greenP}{KPT-encoder}. We then use this prior to initialize the latent code $\vecb{z}_0$ and evaluate the likelihood of the latent code $\vecb{z}_t$ where $t$ corresponds to the optimization step. 
The red dashed lines denote the gradients of the respective objective terms, namely the latent likelihood $\mathcal{L}_L$ (\Eq{flow}) and the reconstruction loss $\mathcal{L}_R$. 
}
\label{fig:model}
\vspace{0.5cm}
\end{figure*} 

\section{Method}
\label{sec:method}
\vspace{-0.5cm}
Our work extends the Deep Appearance Model (DAM) \cite{lombardi2018deep} by introducing a run-time encoder (LiP-encoder) that is conditioned on the impoverished inference-time inputs. 
We aim to learn models that are capable of driving avatars via partial HMC images or sparse facial landmarks. 
Our paper showcases two modalities for different yet related tasks, including but not limited to HMC- and KPT-encoder taking HMC views (see \Fig{model}) and 2D keypoints, respectively. 
In the remainder of the paper, the LiP-encoder term refers to either of the HMC- or KPT-encoder. 
We treat a LiP-encoder as a conditional prior and train it end-to-end together with the DAM network via our flow-based latent space formulation.

In the following sections, we first explain the problem setup and provide an overview of the base 3D face model, DAM \cite{lombardi2018deep}, in \Sec{problem_setup}. We then present our approach, \model, in \Sec{lip_flow} along with the HMC- and KPT-encoder networks in \Sec{lip_encoders}. We also describe alternative techniques to drive the avatar with inference-time inputs in \Sec{experiments}. Architecture details are in the \Supp

\vspace{-0.2cm}
\subsection{Background: Deep Appearance Model}
\label{sec:problem_setup}
\vspace{-0.1cm}
We follow the data preprocessing steps described in \cite{lombardi2018deep} and use the same data, and training routines for all the models. Specifically, frames are unwarped into a texture, using the tracked geometry. The DAM-encoder ($Q$) (\Fig{overview}-a) takes the geometry \small$\vecb{M} \in \real^{7306 \times 3}$\normalsize and the average texture across all the views \small$\bar{\vecb{T}} \in \real^{3 \times 1024 \times 1024}$\normalsize, and parameterizes a Normal distribution of latent codes \small$\vecb{z} \in \real^{256}$\normalsize:
\begin{small}
\begin{equation}
    \bmQ{\mu}, \bmQ{\sigma} = Q(\bar{\vecb{T}}, \vecb{M}), \quad \vecb{z} \sim \gaussian(\bmQ{\mu}, \bmQ{\sigma}).
    \label{eq:q_encoding}
\end{equation}
\end{small}
The decoder $D$ decodes a latent sample $\vecb{z}$ for a given camera view-vector $\vecb{v}$ into geometry $\hat{\vecb{M}}$ and view-specific texture $\bar{\vecb{T}}^v$ accounting for view-dependent effects by $\hat{\vecb{M}}, \bar{\vecb{T}}^v = D(\vecb{z}, \vecb{v})$. The reconstructed image is rendered with the decoded texture, mesh, and camera pose information. The training objective consists of reconstruction and latent space regularization terms with separate weights:
\begin{small}
\begin{align}
    \mathcal{L} &= \sum_{v} \lambda_{I}\mathcal{L}_{I} + \lambda_{M}\mathcal{L}_{M} + \lambda_{L}\mathcal{L}_{L}, 
    \label{eq:training_objective} \\
    \mathcal{L}_{I} &= \norm{\left(\vecb{I}^v - \hat{\vecb{I}}^v\right)\odot \vecb{W}^v}^2, 
    \quad
    \mathcal{L}_{M} = \norm{\vecb{M} - \hat{\vecb{M}}}^2, \\ 
    \mathcal{L}_{L} &= D_{KL}\left(\gaussian(\bmQ{\mu}, \bmQ{\sigma})~ \Vert ~\gaussian(\vecb{0}, \vecb{I})\right),
    \label{eq:kl_term}
\end{align}
\end{small}

where $\mathcal{L}_{I}$ is the image reconstruction loss between the rendered $\hat{\vecb{I}}^v$ and the ground-truth image \small$\vecb{I}^v \in \real^{3\times 1334 \times 2048}$\normalsize, given view $\vecb{v}$. The view-dependent image masks $\vecb{W}^v$ removes the background. We apply additional supervision for the predicted mesh via a geometry reconstruction loss $\mathcal{L}_{M}$. DAM uses a standard Gaussian prior (\Fig{overview}-a) where its latent space is regularized via KL-D ($\mathcal{L}_{L}$). This formulation already achieves high-quality reconstructions during training. However, we show that the prior $\gaussianpri$ is not effective enough when recovering underlying latent codes only from the impoverished observations (see \Fig{iterative_fitting}-c).
for the latent code fitting task with only partial observations (see \Fig{iterative_fitting}-c).

\vspace{-0.2cm}
\subsection{Inference-time Encoders (LiP-encoder)}
\label{sec:lip_encoders}
\begin{figure*}[t]
\begin{center}
\includegraphics[width=\linewidth]{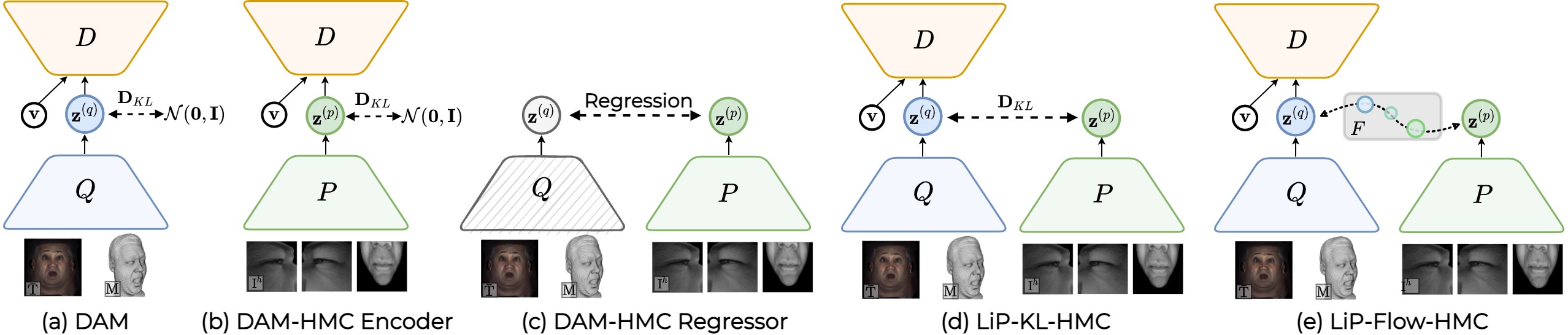}
\end{center}
\vspace{-0.4cm}
\caption{
\textbf{An overview of latent space concepts.} Note that the models are depicted with HMC inputs. The same concepts also apply to the KPT-encoder with keypoint inputs. Details are skipped for brevity.
The decoder $D$ decodes a view-specific texture and mesh for a given view vector $\vecb{v}$ and a latent code $\vecb{z}$. (a) The Deep Appearance Model (DAM) following the conditional VAE framework with $\gaussian(0,I)$ prior. (b) Replacing the DAM-encoder $Q$ with the HMC-encoder $P$. The face avatar model (i.e., the decoder) is trained with the HMC images directly. (c) Learning to regress the latent codes for HMC views by using a pretrained DAM-encoder $Q$. (d) Training the HMC-encoder as a conditional prior model by minimizing the KL-divergence objective. (e) Our \modelname introduces a normalizing flow bridging the latent space of $Q$ and $P$.
}
\label{fig:overview}
\vspace{0.2cm}
\end{figure*}

We define the HMC- and the KPT-encoders conditioned on the HMC images and the 2D keypoints, respectively. Note that these networks are also used in the baselines illustrated in \Fig{overview}-b and \Fig{overview}-c.

\begin{figure}[t]
  \begin{minipage}[c]{0.64\textwidth}
    \includegraphics[width=\textwidth]{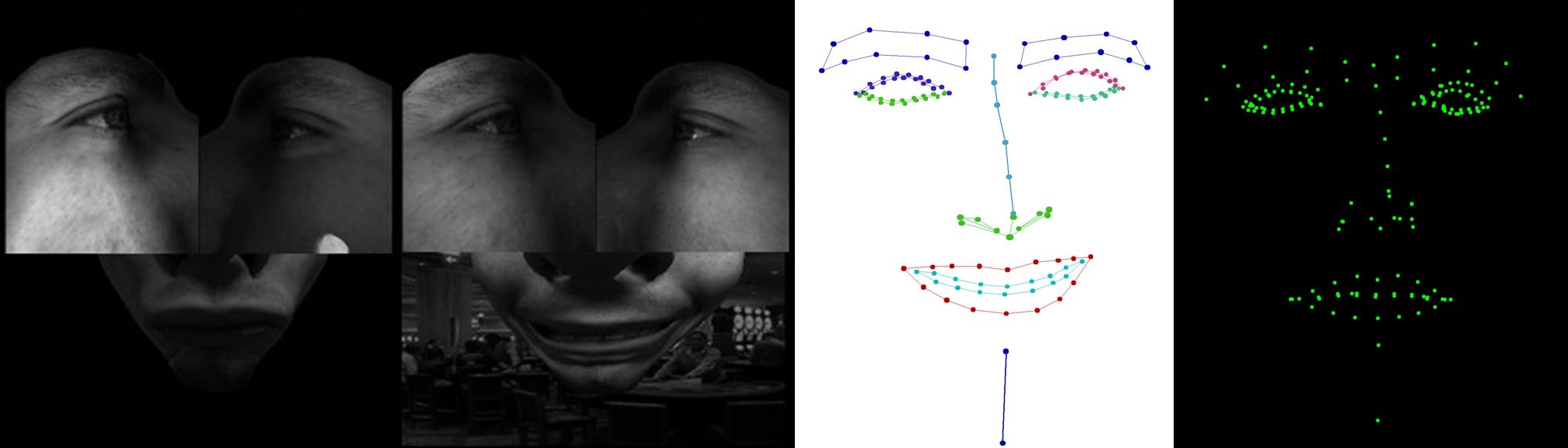}
  \end{minipage}\hfill
  \begin{minipage}{0.34\textwidth}
    \vspace{-10pt}
    \caption{
        \textbf{Driving signals.} Synthetic HMC images with lighting and background augmentations (left) and 2D keypoints (right).
    }
    \label{fig:hmc_imgs}
  \end{minipage}
  \vspace{0.2cm}
\end{figure}

\vspace{-0.2cm}
\paragraph{HMC-encoder} The set of HMC views at every frame consists of partial and non-overlapping images of mouth $\vecb{H}^m$, left eye $\vecb{H}^l$ and right eye $\vecb{H}^r$ regions (\Fig{hmc_imgs}) where $\vecb{H}^* \in \real^{480 \times 640}$. 
The HMC-encoder consists of small networks for each view which are then fused via a $3$-channel attention operation:
\begin{small}
\begin{equation}
    \vecb{h}^* = P^*(\vecb{H}^*), \quad \quad \vecb{h} = \text{Attention}(\vecb{h}^m, \vecb{h}^l, \vecb{h}^r; \vecb{W}^A),
    \label{eq:hmc_encoder_h}
\end{equation}
\end{small}
where $\vecb{h}^*$ denotes the representations for each HMC view. The hidden representation and the attention weights are of shape $\vecb{h} \in \real^{1024}$ and $\vecb{W}^A \in \real^{1024 \times 3}$, respectively. The attention weights are data-agnostic trainable parameters.
The attention operation learns a weighted mixture of HMC views.

\vspace{-0.2cm}
\paragraph{KPT-encoder} 2D keypoint samples, $\vecb{K}$, are in pixel coordinates normalized between $0$ and $1$, consisting of facial landmarks except the iris (\Fig{hmc_imgs}) where $\vecb{K} \in \real^{136 \times 2}$. We pass the keypoint samples to a ResNet \cite{he2016deep} with fully connected layers. Our KPT-encoder maps the keypoint inputs to a an over-parameterized representation $\vecb{h} \in \real^{1024}$.

\vspace{-0.2cm}
\paragraph{Latent Space} We model a probabilistic latent space and parameterize a Gaussian distribution as a function of the run-time input context $\vecb{h}$:
\begin{small}
\begin{equation}
    \bmP{\mu}, \bmP{\sigma} = P(\vecb{h}), \quad \vecP{z} \sim \gaussian(\bmP{\mu}, \bmP{\sigma}).
\label{eq:p_encoding}
\end{equation}
\end{small}
Without loss of generality, $P$ encapsulates the HMC- and KPT-encoders and the layers that compute $\bmP{\mu}$ and $\bmP{\sigma}$.

\vspace{-0.3cm}
\subsection{Learning Conditional Priors}
\label{sec:lip_flow}
Instead of a standard Gaussian prior, we propose a learned prior, which has been shown to be more expressive \cite{chung2015recurrent,sohn2015learning}. Here, the prior model is conditioned on the HMC images or 2D keypoints. We also study several architectures for this task, illustrated in \Fig{overview} and explained in \Sec{experiments}. 

\vspace{-0.15cm}
\paragraph{LiP-KL (\Fig{overview}-d)}
In previous studies \cite{chung2015recurrent,sohn2015learning,aksan2019stcn,rempe2021humor}, conditional priors replace the VAE's isotropic Gaussian prior with a learned prior. Similarly in our work, the latent objective $\mathcal{L}_{L}$ in \Eq{kl_term} takes the form of 
\begin{small}
\begin{equation}
    \mathcal{L}_{L} = D_{KL}\left(\gaussian(\bmQ{\mu}, \bmQ{\sigma})~ \Vert ~\gaussian(\bmP{\mu}, \bmP{\sigma})\right), 
    \label{eq:kl_term_cond}
\end{equation}
\end{small}
where the KL-D objective $D_{KL}$ enforces similarity between $Q$ and $P$. We build a variant of our model with the KL objective (see \Fig{overview}-d). Ground-truth correspondences between the inputs of the LiP-encoders and the DAM-encoder allow us to learn a mapping between the latent spaces. 
Note that the distributional similarity is a plausible assumption in the prior works where the inputs are often consecutive timesteps.
However, in our case, the discrepancy between the $Q$ and the $P$ inputs is considerably larger than in prior work. 

\vspace{-0.15cm}
\paragraph{LiP-Flow (\Fig{overview}-e)}
To further mitigate the asymmetry of the $Q$ and $P$ spaces, we propose to bridge the two spaces and remove the assumption that they must be the same (\ie, KL-D). To achieve this, we introduce a normalizing flow network $F$ to transform latent samples from one space to another (\Fig{overview}-e). Normalizing flows are typically used to represent complex and intractable data distributions with a simple parametric distribution such as the standard Gaussian \cite{dinh2014nice,rezende2015variational,dinh2016density,kingma2018glow}. In our work, we use a conditional normalizing flow \cite{dinh2014nice,winkler2019learning,bhattacharyya2019conditional} enabling the transformation:
\begin{small}
\begin{equation}
    \bar{\textbf{z}}^{(p)} = F(\vecQ{z}; \vecb{h}), \quad F = f_K \circ \dots \circ f_2 \circ f_1,  
    \label{eq:flow_forward}
\end{equation}
\end{small}
where the bijection $F: Q \rightarrow P$ is a composition of $K$ transformations $f_k$. For a given pair of inference- and training-time inputs, we first draw a $\vecQ{z}$ sample (\Eq{q_encoding}) and estimate a prior distribution by using the LiP-encoder, parameterizing it as a Gaussian (\Eq{p_encoding}). The $\vecQ{z}$ sample is then transformed to the $P$ space corresponding to $\bar{\textbf{z}}^{(p)}$ (\Eq{flow_forward}). Finally, we calculate the log-likelihood of the latent sample $\vecQ{z}$ under the prior distribution $P$ such that:
\begin{small}
\begin{equation}
    \log p_{\scriptscriptstyle P}(\vecQ{z}) = \log p_{\scriptscriptstyle P}(\bar{\textbf{z}}^{(p)}) + \log \left(\text{det} \left\vert \frac{\partial F(\vecQ{z}; \vecb{h})}{\partial\vecQ{z}}\right\vert\right), 
    \label{eq:flow}
\end{equation}
\end{small}
where $\log p_{\scriptscriptstyle P}(\bar{\textbf{z}}^{(p)}) = \log \gaussian(\bar{\textbf{z}}^{(p)}; \bmP{\mu}, \bmP{\sigma})$ (see \Eq{p_encoding}). 

In \model, the latent training objective $\mathcal{L}_{L}$ then becomes the negative log-likelihood $-\log p_{\scriptscriptstyle P}(\vecQ{z})$ in \Eq{flow}, replacing the $D_{KL}$ term.
We use the flow $F$ to project the sample $\vecQ{z}$ onto the $P$ space. Hence, this objective enforces $\vecP{z}$ and $\bar{\textbf{z}}^{(p)}$ to be similar \emph{after} the transformation of $\vecQ{z}$ via $F$ (\Eq{flow_forward}). However, the formulation does not restrict $Q$ or $P$ to any particular structure. 

We would like to note that this likelihood can increase arbitrarily by contracting the $P$ distribution, resulting in a trivial solution. This behavior is prevented by the determinant of the Jacobian term in \Eq{flow}, penalizing the contraction and encouraging expansion \cite{dinh2014nice}.

To keep the modifications to the base model at a minimum, the decoder $D$ is trained with samples from the DAM-encoder $Q$ only, as in the original pipeline. We do not use any samples from the LiP-encoders $P$. In other words, the LiP-encoders learn to complete the partial observations via the latent objective (\Eq{flow}). Finally, to attain the reconstruction signal in our \model's latent flow network $F$, the decoder's latent code $\vecb{z}$ follows the flow cycle: $\vecb{z}=F^{-1}(F(\vecQ{z}; \vecb{h}); \vecb{h})$. We provide ablations in \Sec{ablations}.

\vspace{0.2cm}
\section{Results}
\label{sec:experiments}
\vspace{-0.2cm}
\noindent We experimentally answer the following questions: (1) How does end-to-end training of the conditional priors affect the base model's performance? (2) What is the most effective approach of leveraging limited driving signals?
(3) Which of the priors is the most useful in inference-time optimization tasks? 

The base DAM network consists of a view-conditioned decoder and an encoder expecting rich training data. To evaluate the DAM in run-time conditions, we follow analysis-by-synthesis and optimize latent codes to reconstruct impoverished inference-time observations. For a fair comparison, we also introduce the HMC- or the KPT-encoders into the DAM, leveraging the run-time inputs in different ways and allowing us to make latent code predictions directly.

\begin{figure*}[t]
\begin{center}
\includegraphics[width=\linewidth, trim={0 840 280 0},clip]{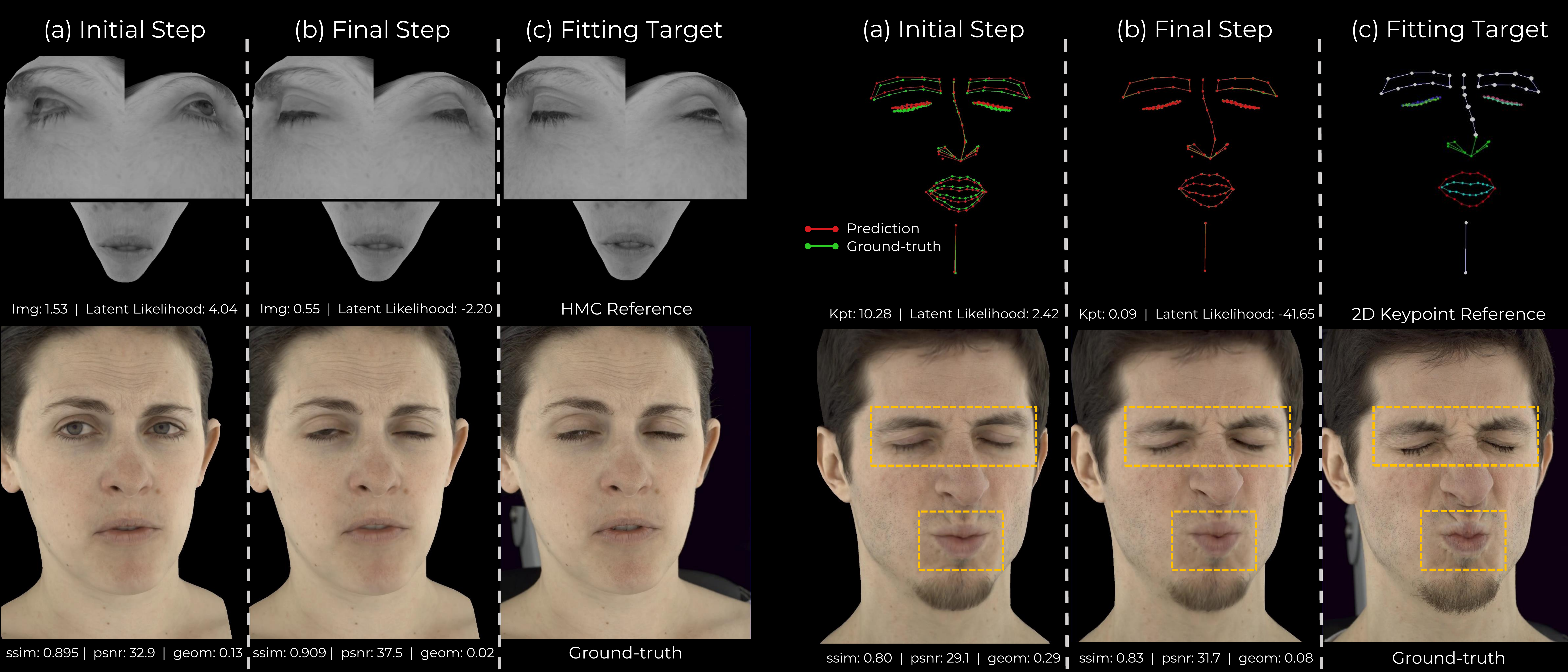}
\end{center}
\vspace{-0.4cm}
\caption{\textbf{Reconstruction from inference-time observations with \model}. We optimize a latent code to reconstruct the 3D avatar (bottom) from imperfect fitting targets (c), such as HMC images (left) or 2D keypoints (right). The latent code is initialized with a sample from the HMC-encoder's or the KPT-encoder's prior. 
We provide the renderings from the initial latent code in columns (a) both in the target's representation and in the frontal view.
The optimization considers the fitting loss between the targets and the HMC (\ie ``Img'' loss) or 2D keypoint (\ie ``Kpt'' loss) projections (\Eq{fitting_img}), and the latent likelihood under the corresponding prior model (\Eq{fitting_logli}). Renderings after the optimization are in the ``Final Step'' columns (b). 
}
\label{fig:fitting_hmc_subj2}
\vspace{0.4cm}
\end{figure*}

\vspace{-0.3cm}
\paragraph{DAM with an inference-time encoder (\Fig{overview}-b)}
The naive way to attain a face avatar from the inference-time inputs is to use the ground-truth correspondences between the multi-view images and the synthetic HMC data or 2D keypoints. To do so, we replace the DAM-encoder with an HMC- or a KPT-encoder and train a conditional VAE as before. We use this model to verify the hypothesis that the limited run-time observations do not carry enough information to faithfully build a high-quality avatar model.

\vspace{-0.3cm}
\paragraph{Inference-time encoder as regressor (\Fig{overview}-c)}
To analyse the effectiveness of treating the inference-time encoders as a prior and inference-time optimization, we train the KPT- and HMC-encoders to regress the latent codes of the pre-trained DAM-encoder.
We follow the probabilistic approach in \Eq{p_encoding} and use the negative log-likelihood training objective.

In our work, the latent space formulation is the only difference across models (\Fig{overview}). To tightly control the settings and simulate the application conditions, we introduce facial landmark and synthetic HMC datasets (\Fig{hmc_imgs}). The synthetic HMC images are generated by re-projecting the multi-view camera images into virtual head-mounted camera views with various lighting, background and headset slop augmentations. For the 2D keypoints, we project a predetermined set of mesh vertices onto the image plane. Both the 2D keypoints and synthetic HMC images provide correspondences to ground truth 3D full face observations, enabling careful study of the methods. We report average performance over $4$ subjects in the PSNR and SSIM image metrics, and per vertex geometry loss.

In \Sec{reconstruction}, we present reconstruction performance of the models in the training setup.
We then compare our model \modelname against the baselines via inference-time optimization with limited observations in \Sec{fitting_overview}. We report the performance of the models with both the HMC-encoder and KPT-encoder as the prior model. Finally, we provide insights on dynamics of our model in \Sec{latent_space}. 
In our \Supp, we provide experiment details, additional evaluations (\Sec{fitting_ablations}), qualitative results (\Sec{sup_qual_results}) and an ablation (\Sec{uncond_flow}) where we use our flow-based formulation with a $\gaussianpri$ instead of a conditional prior.

\begin{table*}[t]
\vspace{-0.3cm}
\begin{minipage}[c]{0.5\textwidth}
\renewcommand{\arraystretch}{1.4}
\setlength\tabcolsep{7pt}%
\resizebox{\linewidth}{!}{%
\small
\begin{NiceTabular}{lccc}[colortbl-like]
& \multicolumn{3}{c}{\textbf{Decoding DAM-encoder ($Q$)}} \\
\multicolumn{1}{l}{\textbf{Models}} & \textbf{PSNR $\uparrow$} & \textbf{SSIM $\uparrow$} & \textbf{Geom $\downarrow$} \\ \hline
\rowcolor[HTML]{FFFFFF} 
\cellcolor[HTML]{FFFFFF}\textbf{(a)} \textbf{DAM} & 36.05 & 0.893 & 0.015 \\ \hline
\rowcolor[HTML]{EFEFEF} 
\cellcolor[HTML]{EFEFEF}\textbf{(b)} \textbf{DAM-HMC Enc.} & 33.93 & 0.872 & 0.186 \\
\rowcolor[HTML]{FFFFFF} 
\cellcolor[HTML]{FFFFFF}\textbf{(c)} \textbf{DAM-HMC Reg.} & n/a & n/a & n/a \\
\rowcolor[HTML]{EFEFEF} 
\cellcolor[HTML]{EFEFEF}\textbf{(d)} \textbf{LiP-KL-HMC} & 35.79 & 0.890 & 0.018 \\
\rowcolor[HTML]{FFFFFF} 
\cellcolor[HTML]{FFFFFF}\textbf{(e)} \textbf{LiP-Flow-HMC} & \textbf{36.21} & \textbf{0.895} & \textbf{0.014} \\ \hline
\rowcolor[HTML]{EFEFEF} 
\textbf{(b)} \textbf{DAM-KPT Enc.} & 33.50 & 0.866 & 0.142 \\
\rowcolor[HTML]{FFFFFF} 
\textbf{(c)} \textbf{DAM-KPT Reg.} & n/a & n/a & n/a \\
\rowcolor[HTML]{EFEFEF} 
\textbf{(d)} \textbf{LiP-KL-KPT} & 35.71 & 0.888 & 0.021 \\
\rowcolor[HTML]{FFFFFF} 
\textbf{(e)} \textbf{LiP-Flow-KPT} & \textbf{36.22} & \textbf{0.895} & \textbf{0.014} \\ \hline
\end{NiceTabular}
}
\end{minipage}\hfill
\begin{minipage}{0.48\textwidth}
\vspace{0.4cm}
\caption{\textbf{Reconstruction.} 
We evaluate the DAM-encoder ($Q$) and the decoder ($D$) by reconstructing the latent codes estimated from the training-time inputs to see how the inference-time encoder affects the base model, DAM.
We report the performance with both the HMC-encoder and the KPT-encoder (\Fig{overview}). The DAM-HMC Enc. and the DAM-KPT Enc. baselines (b) take HMC images and 2D keypoints as inputs, respectively.
}
\label{tab:table_reconstruction}
\end{minipage}
\vspace{-0.1cm}
\end{table*}

\vspace{-0.2cm}
\subsection{Reconstruction Quality}
\label{sec:reconstruction}
\vspace{-0.1cm}
We evaluate the reconstruction performance of the 3D face model in the training setup.
More specifically, the decoder ($D$) reconstructs the latent codes from the DAM-Encoder ($Q$) where we pass the average texture and the 3D geometry (\Eq{q_encoding}). We aim to find the effect of end-to-end training of the conditional priors on the base model.
\Tab{table_reconstruction} provides the reconstruction results when the HMC- and KPT-encoders are treated as priors and trained along with the DAM.

In both setups, our \modelname outperforms the baselines. The \modelKL reduces the reconstruction performance of the base model DAM, suggesting that the combination of conditional prior $(P)$ and the KL objective are detrimental. In contrast, our flow-based prior improves the performance of the DAM, implying that \modelFlow learns a more expressive latent space than the \modelKLnospace. Also note the consistent performance of our flow-based formulation in the HMC- and KPT-encoder settings. Our learned conditional priors tackle the difficulty of modeling the limited inference-time data, and the base model's DAM-encoder ($Q$) and the decoder ($D$) enjoy a more expressive latent space. Since the HMC- and KPT-encoders are required to extract the relevant information only from the limited inputs, the ``DAM-HMC Enc.'' and ``DAM-KPT Enc.'' baselines suffer from underfitting, degrading the performance of the DAM.

% \begin{table*}[t]
% \begin{minipage}[c]{0.5\textwidth}
% \renewcommand{\arraystretch}{1.4}
% \setlength\tabcolsep{7pt}%
% \resizebox{\textwidth}{!}{%
% \small
% \begin{NiceTabular}{b{0.1cm}b{2.8cm}ccc}[colortbl-like]
%  &  & \multicolumn{3}{c}{\textbf{HMC Observations}} \\
% \multicolumn{2}{l}{\textbf{Models}} & \textbf{PSNR $\uparrow$} & \textbf{SSIM $\uparrow$} & \textbf{Geom $\downarrow$} \\ \hline
% \rowcolor[HTML]{FFFFFF} 
% \cellcolor[HTML]{FFFFFF}\textbf{(a)} & \textbf{DAM} & 31.65 & 0.858 & 0.676 \\ \hline
% \rowcolor[HTML]{EFEFEF} 
% \cellcolor[HTML]{EFEFEF}\textbf{(b)} & \textbf{DAM-HMC Enc.} & 33.93 & 0.872 & 0.186 \\
% \rowcolor[HTML]{FFFFFF} 
% \cellcolor[HTML]{FFFFFF}\textbf{(c)} & \textbf{DAM-HMC Reg.} & 34.16 & 0.872 & 0.248 \\ \hline
% \rowcolor[HTML]{EFEFEF} 
% \cellcolor[HTML]{EFEFEF}\textbf{(d)} & \textbf{LiP-KL-HMC} & 34.36 & 0.881 & 0.090 \\
% \rowcolor[HTML]{FFFFFF} 
% \cellcolor[HTML]{FFFFFF}\textbf{(e)} & \textbf{LiP-Flow-HMC} & \textbf{34.98} & \textbf{0.885} & \textbf{0.087} \\ \hline
% \end{NiceTabular}
% }
% \end{minipage}\hfill
% \begin{minipage}{0.48\textwidth}
% \vspace{0.45cm}
% \caption{\textbf{HMC inputs.} We use the HMC-encoder ($P$) to estimate a prior distribution from the HMC images. 
% For the baselines (b) and (c), a latent sample is decoded directly, whereas for (a), (d) and (e), the latent code is first optimized to reconstruct the given HMC images (\Fig{fitting_hmc_subj2}).
% }
% \label{tab:table_hmc_fitting}
% \end{minipage}
% \vspace{-0.2cm}
% \end{table*}

\begin{table*}[t]
\begin{minipage}[c]{0.5\textwidth}
\renewcommand{\arraystretch}{1.4}
\setlength\tabcolsep{7pt}%
\resizebox{\textwidth}{!}{%
\small
\begin{NiceTabular}{lccc}[colortbl-like]
 & \multicolumn{3}{c}{\textbf{HMC Observations}} \\
\multicolumn{1}{l}{\textbf{Models}} & \textbf{PSNR $\uparrow$} & \textbf{SSIM $\uparrow$} & \textbf{Geom $\downarrow$} \\ \hline
\rowcolor[HTML]{FFFFFF} 
\cellcolor[HTML]{FFFFFF}\textbf{(a)} \textbf{DAM} & 31.65 & 0.858 & 0.676 \\ \hline
\rowcolor[HTML]{EFEFEF} 
\cellcolor[HTML]{EFEFEF}\textbf{(b)} \textbf{DAM-HMC Enc.} & 33.93 & 0.872 & 0.186 \\
\rowcolor[HTML]{FFFFFF} 
\cellcolor[HTML]{FFFFFF}\textbf{(c)} \textbf{DAM-HMC Reg.} & 34.16 & 0.872 & 0.248 \\ \hline
\rowcolor[HTML]{EFEFEF} 
\cellcolor[HTML]{EFEFEF}\textbf{(d)} \textbf{LiP-KL-HMC} & 34.36 & 0.881 & 0.090 \\
\rowcolor[HTML]{FFFFFF} 
\cellcolor[HTML]{FFFFFF}\textbf{(e)} \textbf{LiP-Flow-HMC} & \textbf{34.98} & \textbf{0.885} & \textbf{0.087} \\ \hline
\end{NiceTabular}
}
\end{minipage}\hfill
\begin{minipage}{0.48\textwidth}
\vspace{0.45cm}
\caption{\textbf{HMC inputs.} We use the HMC-encoder ($P$) to estimate a prior distribution from the HMC images. 
For the baselines (b) and (c), a latent sample is decoded directly, whereas for (a), (d) and (e), the latent code is first optimized to reconstruct the given HMC images (\Fig{fitting_hmc_subj2}).
}
\label{tab:table_hmc_fitting}
\end{minipage}
\vspace{-0.2cm}
\end{table*}
% \begin{table*}[t]
% \vspace{-0.2cm}
% \begin{minipage}[c]{0.5\textwidth}
% \renewcommand{\arraystretch}{1.4}
% \setlength\tabcolsep{7pt}%
% \resizebox{\linewidth}{!}{%
% \small
% \begin{NiceTabular}{b{0.1cm}b{2.8cm}ccc}[colortbl-like]
%  &  & \multicolumn{3}{c}{\textbf{Keypoint (KPT) Observations}} \\
% \multicolumn{2}{l}{\textbf{Models}} & \textbf{PSNR $\uparrow$} & \textbf{SSIM $\uparrow$} & \textbf{Geom $\downarrow$} \\ \hline
% \rowcolor[HTML]{FFFFFF} 
% \cellcolor[HTML]{FFFFFF}\textbf{(a)} & \textbf{DAM} & 31.62 & 0.865 & 0.431 \\ \hline
% \rowcolor[HTML]{EFEFEF} 
% \cellcolor[HTML]{EFEFEF}\textbf{(b)} & \textbf{DAM-KPT Enc.} & 33.50 & 0.866 & 0.142 \\
% \rowcolor[HTML]{FFFFFF} 
% \cellcolor[HTML]{FFFFFF}\textbf{(c)} & \textbf{DAM-KPT Reg.} & 33.67 & 0.864 & 0.151 \\ \hline
% \rowcolor[HTML]{EFEFEF} 
% \cellcolor[HTML]{EFEFEF}\textbf{(d)} & \textbf{LiP-KL-KPT} & 35.08 & 0.886 & 0.089 \\
% \rowcolor[HTML]{FFFFFF} 
% \cellcolor[HTML]{FFFFFF}\textbf{(e)} & \textbf{LiP-Flow-KPT} & \textbf{35.55} & \textbf{0.891} & \textbf{0.053} \\ \hline
% \end{NiceTabular}
% }
% \end{minipage}\hfill
% \begin{minipage}{0.48\textwidth}
% \vspace{0.45cm}
% \caption{\textbf{2D keypoint inputs.} We use the KPT-encoder ($P$) to estimate a prior distribution from the keypoints. For the baselines (b) and (c), a latent sample is decoded directly, whereas for (a), (d) and (e), the latent code is first optimized to reconstruct the given keypoints (\Fig{fitting_hmc_subj2}).
% }
% \label{tab:table_kpt_fitting}
% \end{minipage}
% \vspace{-0.2cm}
% \end{table*}

\begin{table*}[t]
\vspace{-0.2cm}
\begin{minipage}[c]{0.5\textwidth}
\renewcommand{\arraystretch}{1.4}
\setlength\tabcolsep{7pt}%
\resizebox{\linewidth}{!}{%
\small
\begin{NiceTabular}{lccc}[colortbl-like]
  & \multicolumn{3}{c}{\textbf{Keypoint (KPT) Observations}} \\
\multicolumn{1}{l}{\textbf{Models}} & \textbf{PSNR $\uparrow$} & \textbf{SSIM $\uparrow$} & \textbf{Geom $\downarrow$} \\ \hline
\rowcolor[HTML]{FFFFFF} 
\cellcolor[HTML]{FFFFFF}\textbf{(a)} \textbf{DAM} & 31.62 & 0.865 & 0.431 \\ \hline
\rowcolor[HTML]{EFEFEF} 
\cellcolor[HTML]{EFEFEF}\textbf{(b)} \textbf{DAM-KPT Enc.} & 33.50 & 0.866 & 0.142 \\
\rowcolor[HTML]{FFFFFF} 
\cellcolor[HTML]{FFFFFF}\textbf{(c)} \textbf{DAM-KPT Reg.} & 33.67 & 0.864 & 0.151 \\ \hline
\rowcolor[HTML]{EFEFEF} 
\cellcolor[HTML]{EFEFEF}\textbf{(d)} \textbf{LiP-KL-KPT} & 35.08 & 0.886 & 0.089 \\
\rowcolor[HTML]{FFFFFF} 
\cellcolor[HTML]{FFFFFF}\textbf{(e)} \textbf{LiP-Flow-KPT} & \textbf{35.55} & \textbf{0.891} & \textbf{0.053} \\ \hline
\end{NiceTabular}
}
\end{minipage}\hfill
\begin{minipage}{0.48\textwidth}
\vspace{0.45cm}
\caption{\textbf{2D keypoint inputs.} We use the KPT-encoder ($P$) to estimate a prior distribution from the keypoints. For the baselines (b) and (c), a latent sample is decoded directly, whereas for (a), (d) and (e), the latent code is first optimized to reconstruct the given keypoints (\Fig{fitting_hmc_subj2}).
}
\label{tab:table_kpt_fitting}
\end{minipage}
\vspace{-0.2cm}
\end{table*}

\vspace{-0.3cm}
\subsection{Inference-time Optimization}
\vspace{-0.1cm}
\label{sec:fitting_overview}
Here, we analyze the models in the inference-time conditions. We treat the pre-trained decoder as a black box renderer for a given latent code, view direction, and camera pose and fit the latent code to reconstruct the given target:
\begin{small}
\begin{align}
    &\argmin_{\vecb{z}} \mathcal{L}_{R} + \lambda_L\mathcal{L}_{L} \label{eq:fitting_objective}, \\
    &\mathcal{L}_{R} = \norm{\left(\vecb{I}^v - \hat{\vecb{I}}^v\right)\odot \vecb{W}^v}^2, 
    \label{eq:fitting_img} \\
    &\mathcal{L}_{L} = -\log p_{\scriptscriptstyle P}(\textbf{z}) = -\log \gaussian(\textbf{z}; \bmP{\mu}, \bmP{\sigma}),
    \label{eq:fitting_logli}
\end{align}
\end{small}
where $\bmP{\mu}$ and $\bmP{\sigma}$ are predicted by the HMC- or KPT-encoders. The learned prior is leveraged via the latent likelihood term $\mathcal{L}_{L}$ (\Eq{fitting_logli}). Similarly, the latent codes are initialized with the predicted mean $\bmP{\mu}$. Note that for the base model DAM (\Fig{overview}-a), the prior is a standard Gaussian. 

For all models, we assume that the decoder's input $\textbf{z}$ stems from $Q$. Hence, in \model, the initial latent sample is mapped to the $Q$ space via the inverse of \Eq{flow_forward}. The likelihood is computed via \Eq{flow_forward} and \Eq{flow} (cf. \Fig{model}). 

In this task, we use only the HMC- or KPT-encoders and the decoder ($D$) where we report the optimization performance on the fully rendered images including the parts that are not visible in the fitting targets (see \Fig{fitting_hmc_subj2}).
Fitting results without $\mathcal{L}_L$ is provided in \Sec{fitting_ablations} in the \Supp

\begin{table*}[t]
\vspace{-0.3cm}
\begin{minipage}[c]{0.5\textwidth}
\renewcommand{\arraystretch}{1.4}
\setlength\tabcolsep{7pt}%
\resizebox{\linewidth}{!}{%
\small
\begin{NiceTabular}{lccc}[colortbl-like]
 & \multicolumn{3}{c}{\textbf{Frontal-view Fitting}} \\
\multicolumn{1}{l}{\textbf{Models}} & \textbf{PSNR $\uparrow$} & \textbf{SSIM $\uparrow$} & \textbf{Geom $\downarrow$} \\ \hline
\rowcolor[HTML]{FFFFFF} 
\cellcolor[HTML]{FFFFFF}\textbf{(a)} \textbf{DAM} & 35.16 & 0.888 & 0.212 \\ \hline
\rowcolor[HTML]{EFEFEF} 
\cellcolor[HTML]{EFEFEF}\textbf{(b)} \textbf{DAM-HMC Enc.} & 35.36 & 0.883 & 0.047 \\
\rowcolor[HTML]{FFFFFF} 
\cellcolor[HTML]{FFFFFF}\textbf{(c)} \textbf{DAM-HMC Reg.} & 36.28 & 0.895 & 0.057 \\
\rowcolor[HTML]{EFEFEF} 
\cellcolor[HTML]{EFEFEF}\textbf{(d)} \textbf{LiP-KL-HMC} & 36.08 & 0.892 & 0.039 \\
\rowcolor[HTML]{FFFFFF} 
\cellcolor[HTML]{FFFFFF}\textbf{(e)} \textbf{LiP-Flow-HMC} & \textbf{36.50} & \textbf{0.898} & \textbf{0.022} \\ \hline
\rowcolor[HTML]{EFEFEF} 
\textbf{(b)} \textbf{DAM-KPT Enc.} & 34.75 & 0.874 & 0.059 \\
\rowcolor[HTML]{FFFFFF} 
\textbf{(c)} \textbf{DAM-KPT Reg.} & 36.31 & 0.895 & 0.030 \\
\rowcolor[HTML]{EFEFEF} 
\textbf{(d)} \textbf{LiP-KL-KPT} & 35.94 & 0.890 & 0.041 \\
\rowcolor[HTML]{FFFFFF} 
\textbf{(e)} \textbf{LiP-Flow-KPT} & \textbf{36.53} & \textbf{0.898} & \textbf{0.025} \\ \hline
\end{NiceTabular}
}
\end{minipage}\hfill
\begin{minipage}{0.48\textwidth}
\vspace{0.7cm}
\caption{\textbf{Fitting to frontal-view images.} We evaluate all the models (a-e) in the inference-time optimization task with frontal-view targets.
First, the latent codes are fitted to the frontal-view images and then evaluated by decoding in left, right, and frontal views. Note that the prior input is different from the fitting targets. It is conditioned on the HMC images or 2D keypoints.}
\label{tab:table_frontal_fitting}
\end{minipage}
\vspace{-0.1cm}
\end{table*}

\vspace{-0.2cm}
\paragraph{HMC Targets}
\label{sec:hmc_fitting}
We use the partial HMC views as the fitting target and use the HMC-conditioned priors (see \Fig{fitting_hmc_subj2}-left). We decode the latent code $\vecb{z}$ for $3$ view vectors and render with the corresponding HMC parameters. For each rendered HMC view, the image fitting objective $\mathcal{L}_{R}$ (\Eq{fitting_img}) is calculated on the visible HMC regions only.
This task evaluates the models in terms of multi-view consistency under partial views and the reconstruction quality from incomplete observations. The prior term $\mathcal{L}_L$ (\Eq{fitting_logli}) is of higher importance in this task as the HMC views are unseen and very different from the decoder's training views.

In \Tab{table_hmc_fitting}, our flow-based (e) \modelFlow outperforms the baselines in both the image and geometry metrics. We observe a significant gap between the base model DAM and the models with a learned prior, showing that the standard Gaussian prior is not effective for optimization when only partial information is available. This is partly mitigated by the variants (b) DAM-HMC Enc. and (c) DAM-HMC Reg. which estimate latent codes directly from the observations.

\vspace{-0.2cm}
\paragraph{2D Keypoint Targets}
\label{sec:kpt_fitting}
We evaluate the models by fitting the latent codes to sparse 2D keypoint observations where the priors and the inference-time encoders are also conditioned on the keypoints (see \Fig{fitting_hmc_subj2}-right). After decoding a latent code, a predetermined set of vertices on the predicted mesh is projected onto the image plane. The reconstruction term $L_R$ in \Eq{fitting_img} is the $\ell^2$ norm between the given and the projected keypoints. This means that the latent code is optimized solely based on the geometry and the improved image quality can be attributed to the learned correlations between the texture and the geometry.

The evaluations on keypoints in \Tab{table_kpt_fitting} are inline with the HMC setup (\Tab{table_hmc_fitting}) where our (e) \modelFlow achieves the best performance again, demonstrating the generalization ability of our flow-based latent formulation. Moreover, the geometry error is improved for all the models as the underlying prior model and the fitting objective ($\mathcal{L}_R + \mathcal{L}_L$) explicitly use geometric cues.

\vspace{-0.2cm}
\paragraph{Frontal-view Targets}
\label{sec:frontal_fitting}
In this task, we use frontal-view images as the fitting targets for all the models. 
The models have access to the frontal views targets, and depending on the configuration, the HMC images or the 2D keypoints as the driving signal. This setup allows us to evaluate the models with more informative observations and assess the learned priors when the fitting target is different from the prior inputs. 
\Tab{table_frontal_fitting} summarizes the results for both the HMC- and KPT-encoders as well as the base DAM. Our \modelFlow achieves the best performance in both image and geometry metrics. 
The DAM’s performance significantly improves when the fitting target is more informative frontal views, implying that the DAM is not able to handle impoverished observations.
We also evaluate the DAM baselines (b and c) via latent code fitting. Similar to the DAM, (b) the DAM-HMC Enc. and the DAM-KPT Enc. use a standard Gaussian prior. Our proposed \modelname outperforms the DAM and its variants, showing the advantage of our flow-based conditional prior over the $\gaussianpri$ prior.

\begin{figure}[t]
  \begin{minipage}[c]{0.6\textwidth}
    \includegraphics[width=\textwidth]{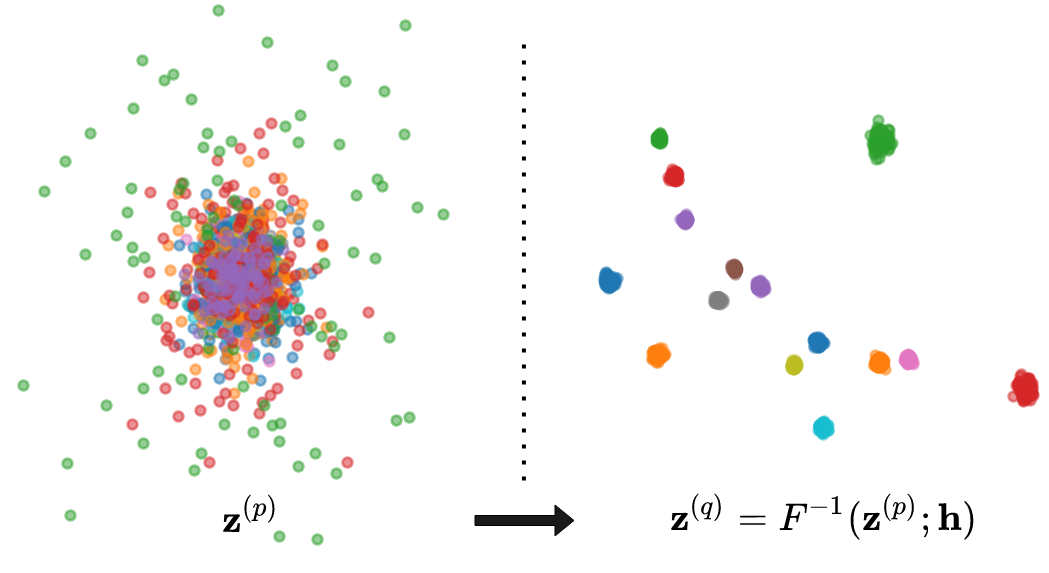}
  \end{minipage}\hfill
  \begin{minipage}{0.38\textwidth}
    \vspace{-10pt}
    \caption{\textbf{Latent space visualization in PCA} (Left) Samples in the $P$ space. (Right) Same samples in the $Q$ space after applying our latent transformation via flow $F$. For a given HMC sample (color-coded), the HMC-encoder predicts a prior distribution. For each of the $15$ HMC inputs, we visualize $100$ latent codes sampled from the respective prior distribution.
    }
    \label{fig:latent_pca_vis}
  \end{minipage}
%   \vspace{-0.1cm}
\end{figure}

\vspace{-0.3cm}
\subsection{Latent Space}
\label{sec:latent_space}
In \Fig{latent_pca_vis}, we visualize the $P$ and $Q$ latent spaces in 2D PCA space. Removing the similarity constraint imposed by the KL-divergence loss results in a highly distorted representation space for the prior model ($P$). It also allows the prior model to assign different amounts of variance to the samples (\ie, the green sample has the largest variance). Considering the performance of our model in the fitting tasks, this unstructured nature of the prior space seems to be effective. In the $Q$ space, however, 
the same latent samples form well-separated clusters for different inputs,
indicating that \modelFlow captures semantics in $Q$.

\vspace{-0.3cm}
\subsection{Ablations}
\label{sec:ablations}
In \Tab{ablation_optim}, we ablate our learned prior in the HMC fitting task. It significantly improves the performance when the latent codes are initialized from the prior and the latent likelihood is considered. Random initializations from the $\gaussianpri$ causes higher geometry error while the latent likelihood $\mathcal{L}_L$ is highly important for image quality. 
We also present two variants of our model. We first train ``LiP-Flow + Decoding $P$'' by decoding samples both from the $P$ and $Q$ spaces, causing a detrimental effect. This setup encourages the LiP-encoders ($P$) to predict the incomplete information in the output space rather than the more compact latent space, limiting the prior's capacity. We then train ``LiP-Flow - Flow Cycle'' by ignoring the flow cycle (cf. \Fig{model}) and decoding $Q$ samples directly. The reconstruction signal helps the latent flow network to learn transformations that are more accurately reflected in the output space.

\begin{table}[t]
\begin{minipage}[c]{0.5\textwidth}
\renewcommand{\arraystretch}{1.4}
\setlength\tabcolsep{7pt}%
\resizebox{\linewidth}{!}{%
\small
\begin{NiceTabular}{l c c c}[colortbl-like]
 & \multicolumn{3}{c}{\textbf{HMC Fitting}} \\
\cellcolor[HTML]{FFFFFF}\textbf{Models}        & PSNR $\uparrow$           & SSIM $\uparrow$  & Geom $\downarrow$ \\ \hline
\textbf{DAM}                           & {31.09} & {0.850} & {0.522} \\ \hline
\rowcolor[HTML]{EFEFEF} 
\textbf{LiP-Flow$-$Flow Cycle}                           & {33.07} & {0.856} & {0.108} \\
\rowcolor[HTML]{FFFFFF} 
\textbf{LiP-Flow$+$Decoding $P$}                           & {33.03} & {0.862} & {0.125} \\
\hline
\rowcolor[HTML]{EFEFEF} 
\textbf{LiP-Flow$-$prior init$-\mathcal{L}_L$} & 28.51 & 0.821 & 0.841 \\
\rowcolor[HTML]{FFFFFF} 
\textbf{LiP-Flow$-$prior init}                   & 32.29 & 0.862 & 0.405   \\
\rowcolor[HTML]{EFEFEF} 
\textbf{LiP-Flow$-\mathcal{L}_L$}         & 30.34& 0.845 & 0.212   \\
\rowcolor[HTML]{FFFFFF} 
\textbf{LiP-Flow}                           & \textbf{33.61} & \textbf{0.874} & \textbf{0.060}
\\ \hline
\end{NiceTabular}
}
%\vspace{0.1cm}
\end{minipage} \hfill
\begin{minipage}[c]{0.5\textwidth}
\vspace{0.7cm}
\caption{\textbf{Ablation.}
(Top) Training our model by ignoring the flow cycle $\vecb{z} = F^{-1}(F(\vecQ{z}; \vecb{h}); \vecb{h})$, and decoding latent samples from the $P$ space as well. 
(Bottom) Evaluating our LiP-Flow by not using initializations from the learned prior and the latent likelihood objective $\mathcal{L}_L$ (\Eq{fitting_logli}). Ablation is performed on one subject.
}
\label{tab:ablation_optim}
\end{minipage}
\end{table}

\vspace{-0.2cm}
\subsection{Discussion}
Our evaluations provide evidence for the effectiveness of our \model. When evaluated with the limited inference-time observations, our proposed conditional prior via our flow-based formulation is superior compared to the standard Normal prior (see Tables \ref{tab:table_hmc_fitting} and \ref{tab:table_kpt_fitting}, and \Fig{iterative_fitting}). This is achieved without sacrificing the base DAM's performance in training-time setup (\Tab{table_reconstruction}). We show that the DAM's performance can be improved by leveraging the inference-time encoders. However, learning an inference-time regressor (DAM-HMC Reg. or DAM-KPT Reg.) separately is not as effective as end-to-end training of the inference-time encoders by treating them as priors. Our \modelFlow yields consistently the best results via inference-time optimization, indicating that our flow-based formulation improves both the prior's and the decoder's performance.
\vspace{-0.4cm}
\section{Limitations and Future Work}
Learning conditional priors relies on the correspondences between the training and inference data, which is not always feasible. For example, in VR telepresence with head-worn displays, we expect a domain gap when attempting to directly replace synthetic HMC images with real ones. Considering the extensive use of synthetic data in other domains \cite{wood2021fake,kocabas2021spec}, we think that introducing domain adaptation techniques \cite{wei2019vr,lombardi2018deep} to our pipeline could be a promising direction to make use of unpaired HMC data. 
Another potential direction is a temporal extension where our flow network considers the latent samples from the previous frame. 
It would also allow us to leverage previously refined latent codes at inference-time.

\vspace{-0.2cm}
\section{Conclusion}
Our work introduces a novel representation learning approach to bridge the information
asymmetry between the training and inference domains for 3D face avatar models. 
We propose end-to-end training of the inference-time models as a prior together with the main avatar model to make it aware of the impoverished driving signals.
Our prior model is tied to the main model via a normalizing flow which learns to map samples from the prior's representation space to the main model's, allowing us to define a latent likelihood objective. We present two related tasks where we augment a 3D face avatar model via a learned prior conditioned on either partial HMC images or sparse 2D keypoints. 
We experimentally show that our formulation yields an expressive and flexible latent space. In tightly controlled evaluations, our model \modelname outperforms the base model as well as a set of carefully designed baselines in reconstruction and various inference-time optimization tasks. Importantly, our approach does not require modifications to the base model or additional training objectives.

%%%%%%%%%% REFERENCES
%{
%	\clearpage
%	\small
%	\bibliographystyle{ieee_fullname}
%	\bibliography{macros,main}
%}

\clearpage
\small
\bibliographystyle{splncs04}
\bibliography{main}

\clearpage
\clearpage

\setcounter{page}{1}

\begin{center}
\large{
Supplementary Material for \\
}
\vspace{0.5em}
\large{
\textbf{LiP-Flow: Learning Inference-time Priors for Codec Avatars via Normalizing Flows in Latent Space} \\
\vspace{1.5em}
}
\vspace{-0.8cm}
\end{center}

\begin{figure*}[h]
\begin{center}
\includegraphics[width=\linewidth]{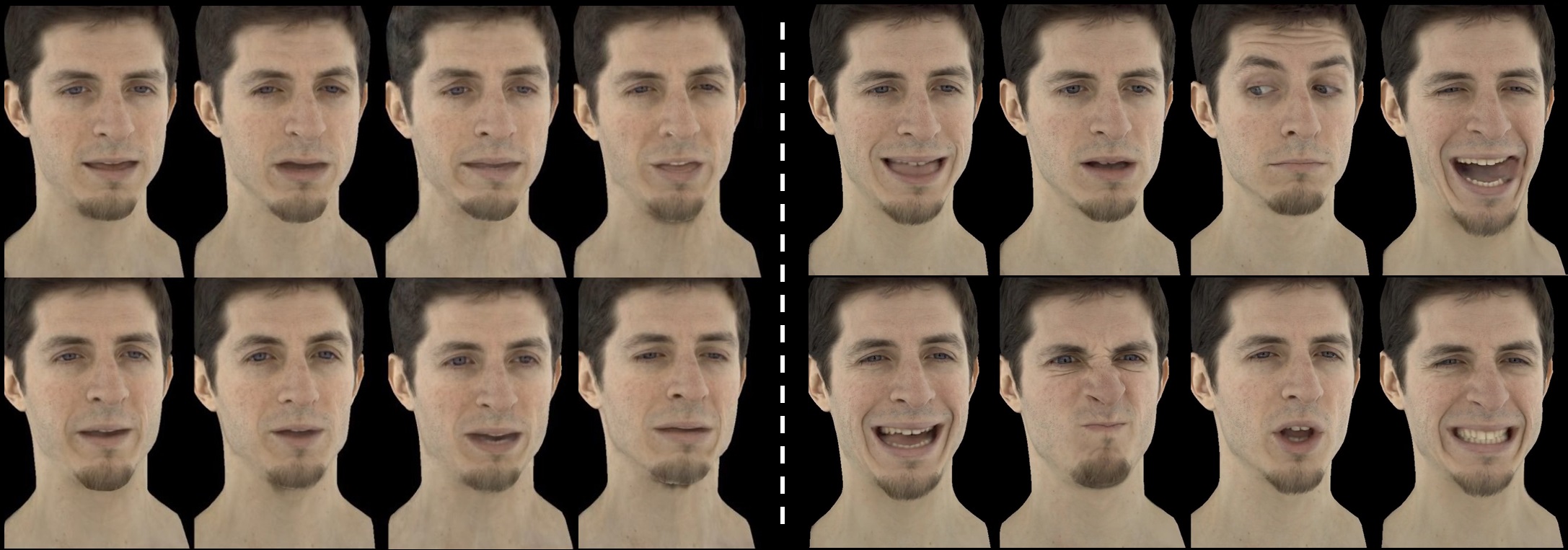}
\end{center}
\vspace{-0.3cm}
    \caption{\textbf{Random latent samples decoded for the frontal view}. (Left) DAM results sampled from $\gaussianpri$. (Right) Our DAM-Flow variant with an unconditional normalizing flow between the DAM-encoder and the standard Normal prior $\gaussianpri$ (\Sec{uncond_flow}). Before passing to the decoder, latent samples are first transformed to the DAM-encoder's (Q) space via $\vecb{z} = F^{-1}(\vecP{z})$. Our flow-based formulation yields a much more expressive latent space, generating higher-quality samples with diverse expressions compared to the KL-based latent space. 
    }
    \label{fig:sup_random_samples}
    \vspace{-0.3cm}
\end{figure*}

This Supplementary Material includes this document and a video. We provide additional results, qualitative evaluations, implementation details of the models and experimental details. Finally, we provide our broader impact statement, in \Sec{broaded_impact}.

\vspace{-0.3cm}
\section{DAM with Unconditional Latent Flow}
\begin{figure}[t]
  \begin{minipage}[c]{0.52\textwidth}
    \includegraphics[width=\textwidth]{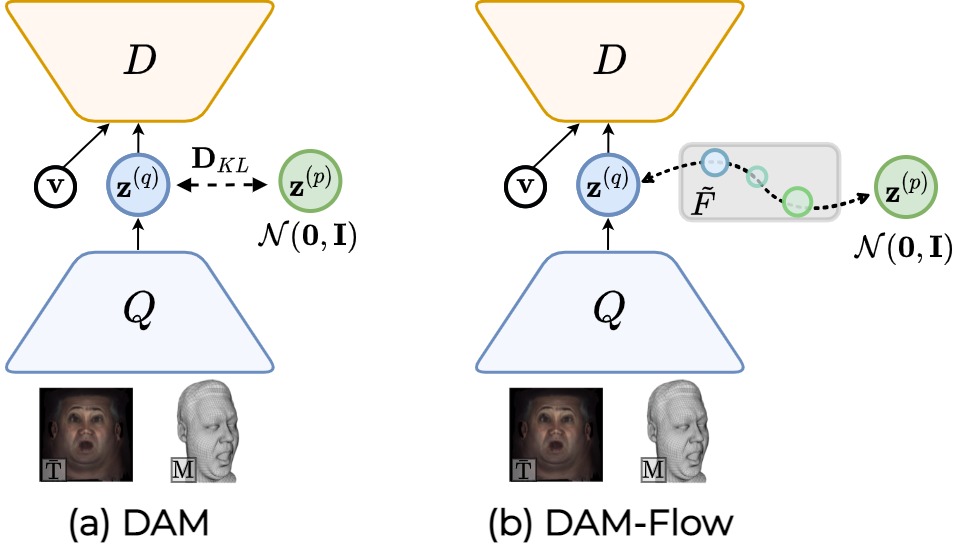}
  \end{minipage}\hfill
  \begin{minipage}{0.44\textwidth}
    \vspace{-10pt}
    \caption{\textbf{DAM with unconditional latent flow}. We replace KL-divergence with our flow-based latent formulation where the prior distribution is a standard Gaussian $\gaussianpri$ in contrast to the learned prior in our main model \modelname.
    }
    \label{fig:sup_dam_flow}
  \end{minipage}
\vspace{0.2cm}
\end{figure}

\vspace{-0.2cm}
\label{sec:uncond_flow}
To evaluate the effectiveness of our flow-based prior formulation, we run an ablation by ignoring the LiP-encoders.
 To this end, we tie a standard Normal prior $\gaussianpri$ to the DAM-encoder ($Q$) via an unconditional normalizing flow and compare this to the DAM trained via KL-D (see \Fig{sup_dam_flow}). We follow the same formulation with \modelname except that we use a standard Normal prior instead of a conditional learned prior and an unconditional normalizing flow model $\tilde{F}$. 

\begin{small}
\begin{align}
    &\bar{\textbf{z}}^{(p)} = \tilde{F}(\vecQ{z}), \quad \tilde{F} = \tilde{f}_K \circ \dots \circ \tilde{f}_2 \circ \tilde{f}_1,  
    \label{eq:sup_flow_forward} \\
    &\log p_{\scriptscriptstyle P}(\vecQ{z}) = \log p_{\scriptscriptstyle P}(\bar{\textbf{z}}^{(p)}) + \log \left(\text{det} \left\vert \frac{\partial \tilde{F}(\vecQ{z})}{\partial\vecQ{z}}\right\vert\right), 
    \label{eq:sup_flow}
\end{align}
\end{small}

\begin{table}[t]
\centering
\begin{minipage}[c]{0.8\textwidth}
\renewcommand{\arraystretch}{1.4}
\setlength\tabcolsep{6pt}%
\resizebox{\linewidth}{!}{%
\small
% Rounded values
\begin{NiceTabular}{lcccccc}[colortbl-like]
% \begin{tabular}{lcccccc}
                  & \multicolumn{3}{c}{\textbf{DAM-encoder Decoding}}                                                      & \multicolumn{3}{c}{\textbf{Frontal Fitting}}          \\
                  & PSNR $\uparrow$ & SSIM $\uparrow$ & \multicolumn{1}{c|}{Geom $\downarrow$}                       & PSNR $\uparrow$ & SSIM $\uparrow$ & Geom $\downarrow$ \\ \hline
\rowcolor[HTML]{FFFFFF} 
\textbf{DAM}      & 34.95           & 0.879          & \multicolumn{1}{c|}{\cellcolor[HTML]{FFFFFF}0.013}          & 34.67           & 0.879          & 0.180            \\
\rowcolor[HTML]{EFEFEF} 
\textbf{DAM-Flow} & 35.02           & 0.881          & \multicolumn{1}{c|}{\cellcolor[HTML]{EFEFEF}\textbf{0.012}} & 35.49           & 0.885          & 0.067            \\ \hline
\rowcolor[HTML]{EFEFEF} 
\textbf{LiP-Flow-HMC}          & 35.20  & \textbf{0.882} & \multicolumn{1}{c|}{\cellcolor[HTML]{EFEFEF}\textbf{0.012}}          & \textbf{35.91}  & \textbf{0.889} & \textbf{0.017}   \\ 
\rowcolor[HTML]{EFEFEF} 
\textbf{LiP-Flow-KPT}          & \textbf{35.25}  & \textbf{0.882}  & \multicolumn{1}{c|}{\cellcolor[HTML]{EFEFEF}\textbf{0.012}}          & 35.90  &\textbf{0.889} &  \textbf{0.017}   \\ \hline
% \end{tabular}
\end{NiceTabular}
}
\end{minipage}\hfill
\begin{minipage}{\textwidth}
\vspace{0.2cm}
\caption{\textbf{DAM-Flow, DAM with unconditional latent flow}. Introducing our flow-based latent space formulation into the base model with a standard Gaussian prior. We report reconstruction performance of the DAM-encoder and the inference-time optimization performance with frontal-view targets on one subject. 
}
\label{tab:dam_flow}
\end{minipage}
\vspace{-0.2cm}
\end{table}

\begin{table*}[t]
\centering
\begin{minipage}[c]{0.9\textwidth}
\renewcommand{\arraystretch}{1.3}
\setlength\tabcolsep{7pt}%
\resizebox{\linewidth}{!}{%
\small
%% Color-coded rows.
\begin{NiceTabular}{lccc:ccc}[colortbl-like]
\textbf{Models} & \textbf{PSNR $\uparrow$} & \textbf{SSIM $\uparrow$} & \textbf{Geom $\downarrow$} & \textbf{PSNR $\uparrow$} & \textbf{SSIM $\uparrow$} & \textbf{Geom $\downarrow$} \\ \hline
 & \multicolumn{3}{c}{\textbf{HMC-Encoder Decoding}} & \multicolumn{3}{c}{\textbf{DAM-Encoder Decoding}} \\
\rowcolor[HTML]{E0E2FF} 
\cellcolor[HTML]{E0E2FF}\textbf{(a) DAM} & n/a & n/a & \cellcolor[HTML]{E0E2FF}n/a & 36.05 & 0.893 & 0.015 \\
\rowcolor[HTML]{FFFFC7} 
\cellcolor[HTML]{FFFFC7}\textbf{(b) DAM-HMC Enc.} & 33.93 & 0.872 & \cellcolor[HTML]{FFFFC7}0.186 & 33.93 & 0.872 & 0.186 \\
\rowcolor[HTML]{AAEFEC} 
\cellcolor[HTML]{AAEFEC}\textbf{(c) DAM-HMC Reg.} & 34.16 & 0.872 & \cellcolor[HTML]{AAEFEC}0.248 & n/a & n/a & n/a \\
\rowcolor[HTML]{FEE1DF} 
\cellcolor[HTML]{FEE1DF}\textbf{(d) LiP-KL-HMC} & 34.39 & 0.876 & \cellcolor[HTML]{FEE1DF}0.146 & 35.79 & 0.890 & 0.018 \\
\rowcolor[HTML]{D0F4DE} 
\cellcolor[HTML]{D0F4DE}\textbf{(e) LiP-Flow-HMC} & 34.04 & 0.871 & \cellcolor[HTML]{D0F4DE}0.161 & \textbf{36.21} & \textbf{0.895} & \textbf{0.014} \\ \hline
 & \multicolumn{3}{c}{\textbf{HMC Fitting w/o $\mathcal{L}_L$}} & \multicolumn{3}{c}{\textbf{HMC Fitting}} \\
\rowcolor[HTML]{E0E2FF} 
\cellcolor[HTML]{E0E2FF}\textbf{(a) DAM} & 31.00 & 0.853 & 0.742 & 31.65 & 0.858 & \cellcolor[HTML]{E0E2FF}0.676 \\
\rowcolor[HTML]{FFFFC7} 
\cellcolor[HTML]{FFFFC7}\textbf{(b) DAM-HMC Enc.} & 33.68 & 0.873 & 0.189 & 34.14 & 0.877 & \cellcolor[HTML]{FFFFC7}0.133 \\
\rowcolor[HTML]{AAEFEC} 
\cellcolor[HTML]{AAEFEC}\textbf{(c) DAM-HMC Reg.} & 34.28 & 0.881 & 0.158 & 34.58 & 0.882 & \cellcolor[HTML]{AAEFEC}0.159 \\
\rowcolor[HTML]{FEE1DF} 
\cellcolor[HTML]{FEE1DF}\textbf{(d) LiP-KL-HMC} & 33.73 & 0.876 & 0.121 & 34.36 & 0.881 & \cellcolor[HTML]{FEE1DF}0.090 \\
\rowcolor[HTML]{D0F4DE} 
\cellcolor[HTML]{D0F4DE}\textbf{(e) LiP-Flow-HMC} & 33.57 & 0.876 & 0.142 & \textbf{34.98} & \textbf{0.885} & \cellcolor[HTML]{D0F4DE}\textbf{0.087} \\ \hline
 & \multicolumn{3}{c}{\textbf{Frontal Fitting w/o $\mathcal{L}_L$}} & \multicolumn{3}{c}{\textbf{Frontal Fitting}} \\
\rowcolor[HTML]{E0E2FF} 
\cellcolor[HTML]{E0E2FF}\textbf{(a) DAM} & 35.32 & 0.888 & \cellcolor[HTML]{E0E2FF}0.312 & 35.16 & 0.888 & \cellcolor[HTML]{E0E2FF}0.212 \\
\rowcolor[HTML]{FFFFC7} 
\cellcolor[HTML]{FFFFC7}\textbf{(b) DAM-HMC Enc.} & 35.21 & 0.882 & \cellcolor[HTML]{FFFFC7}0.047 & 35.36 & 0.883 & 0.047 \\
\rowcolor[HTML]{AAEFEC} 
\cellcolor[HTML]{AAEFEC}\textbf{(c) DAM-HMC Reg.} & 36.20 & 0.894 & \cellcolor[HTML]{AAEFEC}0.049 & 36.28 & 0.895 & 0.057 \\
\rowcolor[HTML]{FEE1DF} 
\cellcolor[HTML]{FEE1DF}\textbf{(d) LiP-KL-HMC} & 35.96 & 0.891 & \cellcolor[HTML]{FEE1DF}0.045 & 36.08 & 0.892 & 0.039 \\
\rowcolor[HTML]{D0F4DE} 
\cellcolor[HTML]{D0F4DE}\textbf{(e) LiP-Flow-HMC} & 36.10 & 0.893 & \cellcolor[HTML]{D0F4DE}0.067 & \textbf{36.50} & \textbf{0.898} & \textbf{0.022} \\ \hline
 & \multicolumn{3}{c}{\textbf{Masked Frontal Fitting w/o $\mathcal{L}_L$}} & \multicolumn{3}{c}{\textbf{Masked Frontal Fitting}} \\
\rowcolor[HTML]{E0E2FF} 
\cellcolor[HTML]{E0E2FF}\textbf{(a) DAM} & 32.30 & 0.867 & \cellcolor[HTML]{E0E2FF}0.770 & 32.38 & 0.867 & \cellcolor[HTML]{E0E2FF}0.604 \\
\rowcolor[HTML]{FFFFC7} 
\cellcolor[HTML]{FFFFC7}\textbf{(b) DAM-HMC Enc.} & 34.37 & 0.876 & 0.108 & 34.59 & 0.878 & \cellcolor[HTML]{FFFFC7}0.095 \\
\rowcolor[HTML]{AAEFEC} 
\cellcolor[HTML]{AAEFEC}\textbf{(c) DAM-HMC Reg.} & 34.89 & 0.883 & 0.152 & 35.17 & 0.886 & \cellcolor[HTML]{AAEFEC}0.149 \\
\rowcolor[HTML]{FEE1DF} 
\cellcolor[HTML]{FEE1DF}\textbf{(d) LiP-KL-HMC} & 34.93 & 0.883 & 0.094 & 35.17 & 0.885 & \cellcolor[HTML]{FEE1DF}0.071 \\
\rowcolor[HTML]{D0F4DE} 
\cellcolor[HTML]{D0F4DE}\textbf{(e) LiP-Flow-HMC} & 35.08 & 0.886 & 0.088 & \textbf{35.44} & \textbf{0.889} & \cellcolor[HTML]{D0F4DE}\textbf{0.060} \\ \hline

\end{NiceTabular}
}
\end{minipage}
\vspace{0.2cm}
\caption{\textbf{Evaluations in HMC setting}. Reporting average performance over $4$ subjects. (Left) We provide the forward-pass performance of the HMC-encoder as well as the inference-time optimization results without using the latent likelihood $\mathcal{L}_L$ (\Eq{fitting_objective}). (Right) Results for the DAM-encoder's forward-pass and inference-time optimization with the latent likelihood. 
In the fitting tasks, we initialize the latent code with the same latent sample we use for evaluating the HMC-encoder. Hence, the ``HMC-encoder Decoding'' results denote the performance before running fitting. 
}
\label{tab:table_sup_avg}
% \vspace{-0.3cm}
\end{table*}

where $\log p_{\scriptscriptstyle P}(\bar{\textbf{z}}^{(p)}) = \log \gaussian(\bar{\textbf{z}}^{(p)}; \vecb{0}, \vecb{I})$ and $\vecQ{z}$ is a latent sample from the DAM-encoder (Q). Our DAM-Flow is conceptually similar to the LSGM \cite{vahdat2021score} in terms of learning an invertible mapping between the encoder space and a standard Normal $\gaussianpri$. Different from \cite{vahdat2021score}, we use a normalizing flow network instead of a continuous diffusion process to learn the mapping, and our encoder ($Q$) still parameterizes a Gaussian. Moreover, we ignore the negative encoder entropy term in the KL decomposition and the KL-D latent regularization term $\mathcal{L}_L$ (\Eq{kl_term}) in the training objective becomes $-\log p_{\scriptscriptstyle P}(\vecQ{z})$. We find this formulation to be more expressive than the base model even with the standard prior. We show that it improves the base model's reconstruction performance both for training and optimization  (see \Tab{dam_flow}). The DAM-Flow's reconstruction performance is on par with the DAM. However, it achieves significantly better reconstructions in the inference-time optimization task when we fit latent codes to single frontal-view images only. We also see that our learned conditional prior is more favorable compared to the standard Gaussian prior.

\begin{table*}[t]
\centering
\begin{minipage}[c]{0.9\textwidth}
\renewcommand{\arraystretch}{1.3}
\setlength\tabcolsep{7.0pt}%
\resizebox{\linewidth}{!}{%
\small
\begin{NiceTabular}{lccc:ccc}
\textbf{Models} & \textbf{PSNR $\uparrow$} & \textbf{SSIM $\uparrow$} & \textbf{Geom $\downarrow$} & \textbf{PSNR $\uparrow$} & \textbf{SSIM $\uparrow$} & \textbf{Geom $\downarrow$} \\ \hline
 & \multicolumn{3}{c}{\textbf{KPT-Encoder Decoding}} & \multicolumn{3}{c}{\textbf{DAM-Encoder Decoding}} \\
\rowcolor[HTML]{E0E2FF} 
\cellcolor[HTML]{E0E2FF}\textbf{(a) DAM} & n/a & n/a & \cellcolor[HTML]{E0E2FF}n/a & 36.05 & 0.893 & 0.015 \\
\rowcolor[HTML]{FFFFC7} 
\cellcolor[HTML]{FFFFC7}\textbf{(b) DAM-KPT Enc.} & 33.50 & 0.866 & \cellcolor[HTML]{FFFFC7}0.142 & 33.50 & 0.866 & 0.142 \\
\rowcolor[HTML]{AAEFEC} 
\cellcolor[HTML]{AAEFEC}\textbf{(c) DAM-KPT Reg.} & 33.67 & 0.864 & \cellcolor[HTML]{AAEFEC}0.151 & n/a & n/a & n/a \\
\rowcolor[HTML]{FEE1DF} 
\cellcolor[HTML]{FEE1DF}\textbf{(d) LiP-KL-KPT} & 34.21 & 0.872 & \cellcolor[HTML]{FEE1DF}0.128 & 35.71 & 0.888 & 0.021 \\
\rowcolor[HTML]{D0F4DE} 
\cellcolor[HTML]{D0F4DE}\textbf{(e) LiP-Flow-KPT} & 33.72 & 0.863 & \cellcolor[HTML]{D0F4DE}0.150 & \textbf{36.22} & \textbf{0.895} & \textbf{0.014} \\ \hline
 & \multicolumn{3}{c}{\textbf{KPT Fitting w/o $\mathcal{L}_L$}} & \multicolumn{3}{c}{\textbf{KPT Fitting}} \\
\rowcolor[HTML]{E0E2FF} 
\cellcolor[HTML]{E0E2FF}\textbf{(a) DAM} & 31.62 & 0.865 & 0.431 & 31.62 & 0.865 & \cellcolor[HTML]{E0E2FF}0.431 \\
\rowcolor[HTML]{FFFFC7} 
\cellcolor[HTML]{FFFFC7}\textbf{(b) DAM-KPT Enc.} & 33.91 & 0.873 & 0.126 & 33.96 & 0.873 & \cellcolor[HTML]{FFFFC7}0.119 \\
\rowcolor[HTML]{AAEFEC} 
\cellcolor[HTML]{AAEFEC}\textbf{(c) DAM-KPT Reg.} & 34.71 & 0.884 & 0.061 & 34.88 & 0.886 & \cellcolor[HTML]{AAEFEC}0.059 \\
\rowcolor[HTML]{FEE1DF} 
\cellcolor[HTML]{FEE1DF}\textbf{(d) LiP-KL-KPT} & 34.74 & 0.883 & 0.099 & 35.08 & 0.886 & \cellcolor[HTML]{FEE1DF}0.089 \\
\rowcolor[HTML]{D0F4DE} 
\cellcolor[HTML]{D0F4DE}\textbf{(e) LiP-Flow-KPT} & 34.79 & 0.884 & 0.063 & \textbf{35.55} & \textbf{0.891} & \cellcolor[HTML]{D0F4DE}\textbf{0.053} \\ \hline
 & \multicolumn{3}{c}{\textbf{Frontal Fitting w/o $\mathcal{L}_L$}} & \multicolumn{3}{c}{\textbf{Frontal Fitting}} \\
\rowcolor[HTML]{E0E2FF} 
\cellcolor[HTML]{E0E2FF}\textbf{(a) DAM} & 35.32 & 0.888 & \cellcolor[HTML]{E0E2FF}0.312 & 35.16 & 0.888 & \cellcolor[HTML]{E0E2FF}0.212 \\
\rowcolor[HTML]{FFFFC7} 
\cellcolor[HTML]{FFFFC7}\textbf{(b) DAM-KPT Enc.} & 34.74 & 0.874 & \cellcolor[HTML]{FFFFC7}0.060 & 34.75 & 0.874 & 0.059 \\
\rowcolor[HTML]{AAEFEC} 
\cellcolor[HTML]{AAEFEC}\textbf{(c) DAM-KPT Reg.} & 36.27 & 0.895 & \cellcolor[HTML]{AAEFEC}0.032 & 36.31 & 0.895 & 0.030 \\
\rowcolor[HTML]{FEE1DF} 
\cellcolor[HTML]{FEE1DF}\textbf{(d) LiP-KL-KPT} & 35.89 & 0.890 & \cellcolor[HTML]{FEE1DF}0.044 & 35.94 & 0.890 & 0.041 \\
\rowcolor[HTML]{D0F4DE} 
\cellcolor[HTML]{D0F4DE}\textbf{(e) LiP-Flow-KPT} & 36.46 & 0.897 & \cellcolor[HTML]{D0F4DE}0.029 & \textbf{36.53} & \textbf{0.898} & \textbf{0.025} \\ \hline
 & \multicolumn{3}{c}{\textbf{Masked Frontal Fitting w/o $\mathcal{L}_L$}} & \multicolumn{3}{c}{\textbf{Masked Frontal Fitting}} \\
\rowcolor[HTML]{E0E2FF} 
\cellcolor[HTML]{E0E2FF}\textbf{(a) DAM} & 32.30 & 0.867 & \cellcolor[HTML]{E0E2FF}0.770 & 32.38 & 0.867 & \cellcolor[HTML]{E0E2FF}0.604 \\
\rowcolor[HTML]{FFFFC7} 
\cellcolor[HTML]{FFFFC7}\textbf{(b) DAM-KPT Enc.} & 33.94 & 0.869 & 0.106 & 34.16 & 0.870 & \cellcolor[HTML]{FFFFC7}0.085 \\
\rowcolor[HTML]{AAEFEC} 
\cellcolor[HTML]{AAEFEC}\textbf{(c) DAM-KPT Reg.} & 35.01 & 0.884 & 0.060 & 35.13 & 0.884 & \cellcolor[HTML]{AAEFEC}0.055 \\
\rowcolor[HTML]{FEE1DF} 
\cellcolor[HTML]{FEE1DF}\textbf{(d) LiP-KL-KPT} & 34.72 & 0.880 & 0.080 & 35.14 & 0.884 & \cellcolor[HTML]{FEE1DF}0.060 \\
\rowcolor[HTML]{D0F4DE} 
\cellcolor[HTML]{D0F4DE}\textbf{(e) LiP-Flow-KPT} & 35.09 & 0.885 & 0.071 & \textbf{35.45} & \textbf{0.887} & \cellcolor[HTML]{D0F4DE}\textbf{0.054} \\ \hline

\end{NiceTabular}
}
\end{minipage}
\vspace{0.2cm}
\caption{\textbf{Evaluations in 2D key point setting}. Reporting average performance over $4$ subjects. (Left) We provide the forward-pass performance of the KPT-encoder as well as the inference-time optimization results without using the latent likelihood $\mathcal{L}_L$ (\Eq{fitting_objective}). 
(Right) Results for the DAM-encoder's forward-pass and inference-time optimization with the latent likelihood. 
In the fitting tasks, we initialize the latent code with the same latent sample we use for evaluating the KPT-encoder. Hence, the ``KPT-encoder Decoding`` results denote the performance before running fitting. 
}
\label{tab:table_sup_kpt_avg}
% \vspace{-0.3cm}
\end{table*}

In \Fig{sup_random_samples}, we present samples generated by the base network DAM and our variant DAM-Flow. We decode random samples from the standard Gaussian prior $\gaussianpri$ for the frontal view. In our variant DAM-Flow, we first transform the latent samples to the $Q$ space via our latent flow function $F$ such that $\vecb{z} = F^{-1}(\vecP{z})$. Our modification to the base DAM network can be considered minimal. Yet our approach yields more diverse and higher-quality samples. This ablation shows that our flow-based latent formulation is an effective approach, and our contribution is not only from learning a conditional prior, both of which constitute a powerful means to solve our problem.

\vspace{-0.3cm}
\section{Inference-time Optimization Ablations}
\vspace{-0.2cm}
\label{sec:fitting_ablations}
In Tables \ref{tab:table_sup_avg} and \ref{tab:table_sup_kpt_avg}, we provide ablations for inference-time optimization and the latent likelihood term $\mathcal{L}_L$ (\Eq{fitting_objective}). We evaluate all the models used in the main paper in both the HMC (\Tab{table_sup_avg}) and the 2D keypoint (\Tab{table_sup_kpt_avg}) settings. The ``HMC-encoder Decoding'' column in \Tab{table_sup_avg} reports the HMC-encoder's forward pass performance. More specifically, given a set of input HMC images, we estimate a prior distribution via the HMC-encoder, and then decode its mean for the frontal-view direction. Note that we use the same latent code to initialize the fitting tasks. Hence, this evaluation provides the performance before optimization. The same evaluation steps also apply to the keypoint setting (\Tab{table_sup_kpt_avg}) where we evaluate the models with the KPT-encoder expecting 2D keypoints. 

\begin{figure}[t]
  \begin{minipage}[c]{0.54\textwidth}
    \includegraphics[width=\textwidth]{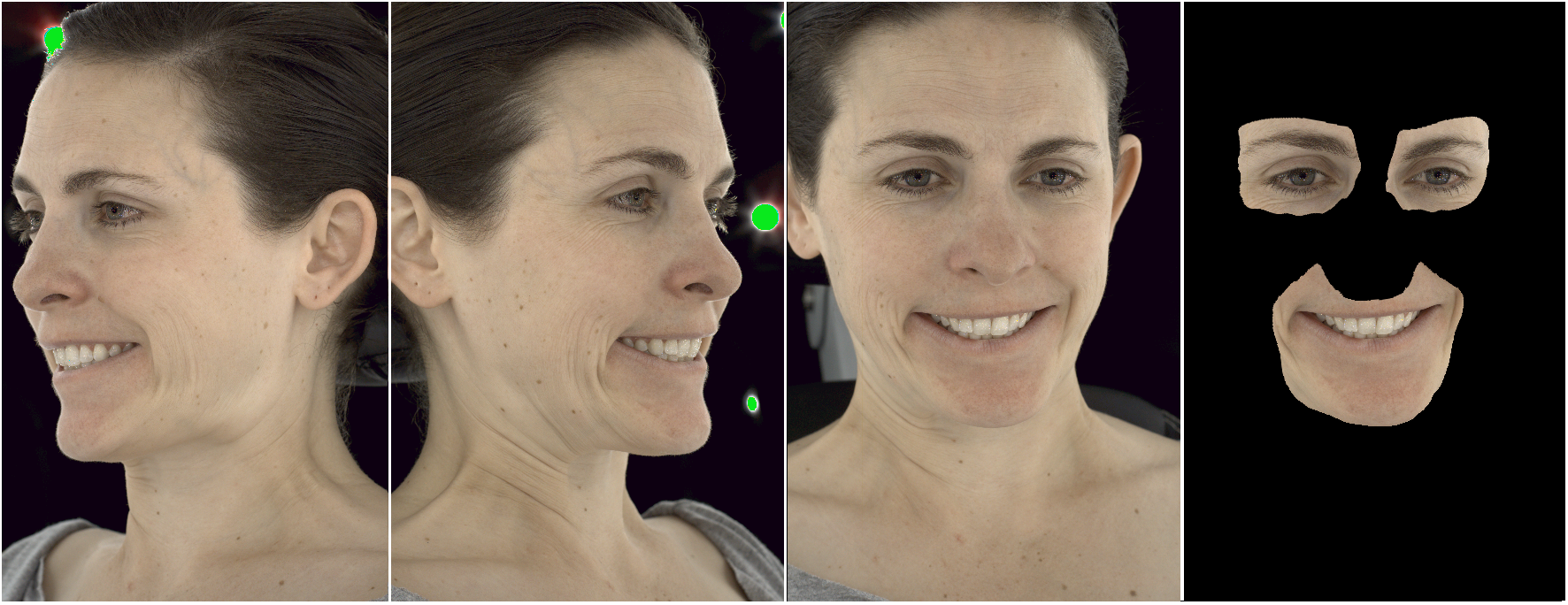}
  \end{minipage}\hfill
  \begin{minipage}{0.44\textwidth}
    \vspace{-10pt}
    \caption{Ground-truth images in left, right, and frontal views. We apply a mask on the frontal-view images to limit the amount of information in the fitting task.
    }
    \label{fig:masked_frontal_targets}
  \end{minipage}
\vspace{0.4cm}
\end{figure}

When we decode the LiP-encoder samples without optimization, we observe that the HMC- and the KPT-encoder trained via our flow-based latent formulation performs poorly whereas the KL-based training achieves the best performance. We think that this is due to the similarity assumption imposed by the KL-divergence loss. It can be explained as a trade-off between the forward-pass accuracy and the iterative fitting performance. While our flow-based formulation yields a prior performing well in the iterative fitting, the KL-based formulation works better in the forward-pass case. However, our proposed LiP-Flow achieves a significant improvement when the latent code is optimized (see ``HMC-encoder Decoding'' vs. ``HMC Fitting'' in \Tab{table_sup_avg} and ``KPT-encoder Decoding'' vs. ``KPT Fitting'' in \Tab{table_sup_kpt_avg}).

In both settings, we also see that using the latent likelihood in the optimization objective is effective for all the models. While its contribution is limited in the baseline models, the learned priors (\ie, Lip-KL and Lip-Flow) benefit the most when the fitting targets carry less information. 

We also provide iterative fitting results for the DAM baselines, namely the ``DAM-HMC Encoder'' and ``DAM-HMC Regressor'' in \Tab{table_sup_avg} and the ``DAM-KPT Encoder'' and ``DAM-KPT Regressor'' in \Tab{table_sup_kpt_avg}. In both settings, both baselines show improvements via iterative fitting. This suggests our proposed inference-time optimization approach is necessary to optimize the performance for all the models.

To make a direct comparison between the clean and noisy targets, we limit the amount of information in the frontal-view targets by applying masks (\Fig{masked_frontal_targets}). We use the same set of target images and mask out the face except the mouth and eye regions. Similar to the ``Frontal-view Fitting'' task (\Tab{table_frontal_fitting}), the models have access to the HMC images or the 2D keypoints as the driving signal depending on the setting and the masked frontal views targets. Although the DAM improves its reconstruction performance when the latent codes are optimized for the frontal-view targets (``Frontal Fitting'' in Tables \ref{tab:table_sup_avg} and \ref{tab:table_sup_kpt_avg}), it degrades significantly on the masked targets. This is inline with the HMC- and keypoint-fitting results.

Finally, in our supplementary video, we provide optimization results with noisy observations where we apply random perturbations to the headset camera parameters. %(starting at 03:00). 
We show that our learned prior is more robust to noise, achieving notably less jitter and more temporally coherent fitting results although we do not apply any temporal regularizations.

\begin{figure*}[t]
\begin{center}
\includegraphics[width=0.9\linewidth]{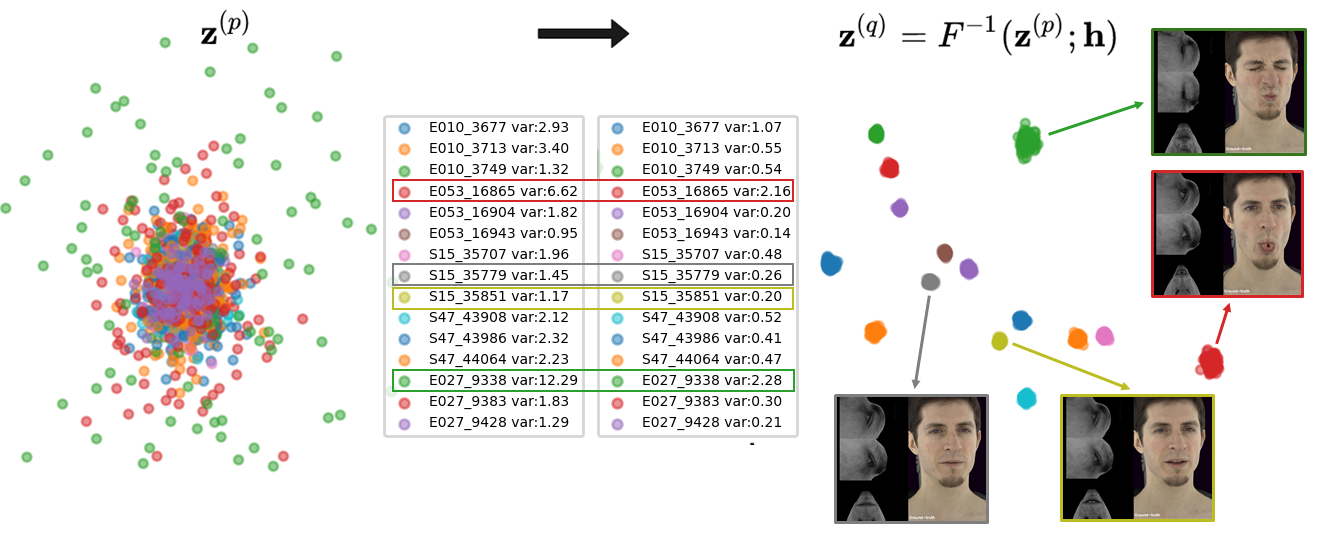}
\end{center}
\vspace{-0.5cm}
\caption{\textbf{Latent space visualization in PCA} (Left) Samples in the $P$ space. (Right) Same samples in the $Q$ space after applying our latent transformation via flow $F$. For a given HMC sample (color-coded), the HMC-encoder $P$ predicts a prior distribution. We visualize $100$ latent codes sampled from the respective prior distribution, in total $1500$ latent samples for $15$ HMC inputs. The legend provides the variance of $100$ latent samples, namely the variance of the predicted prior distribution $\gaussian(\bmP{\mu}, \bmP{\sigma})$ per HMC sample. Samples with neutral expressions result in lower variance whereas rare and peak expressions cause higher variance in the latent space.}
\label{fig:sup_pca}
\vspace{0.4cm}
\end{figure*}

\section{Latent Space}
\label{sec:sup_latent_space}
\vspace{-0.2cm}
\Fig{sup_pca} provides an extended version of the latent space visualization presented in the main submission. We report the variance of the latent samples for every HMC input sample. We also visualize corresponding HMC- and frontal-view images for the samples with high and low variance. We see that our flow-based latent formulation learns to assign larger variance to the inputs with rare and peak facial expressions, quantifying the difficulty of the corresponding sample. 
While the variances are much larger in the prior ($P$) space, they get smaller in the $Q$ space after our latent transformation $F^{-1}$. We provide more results in our supplementary video.

\section{Qualitative Results}
\label{sec:sup_qual_results}
\begin{figure*}[t]
\begin{center}
\includegraphics[width=\linewidth, trim={0 450 460 0},clip]{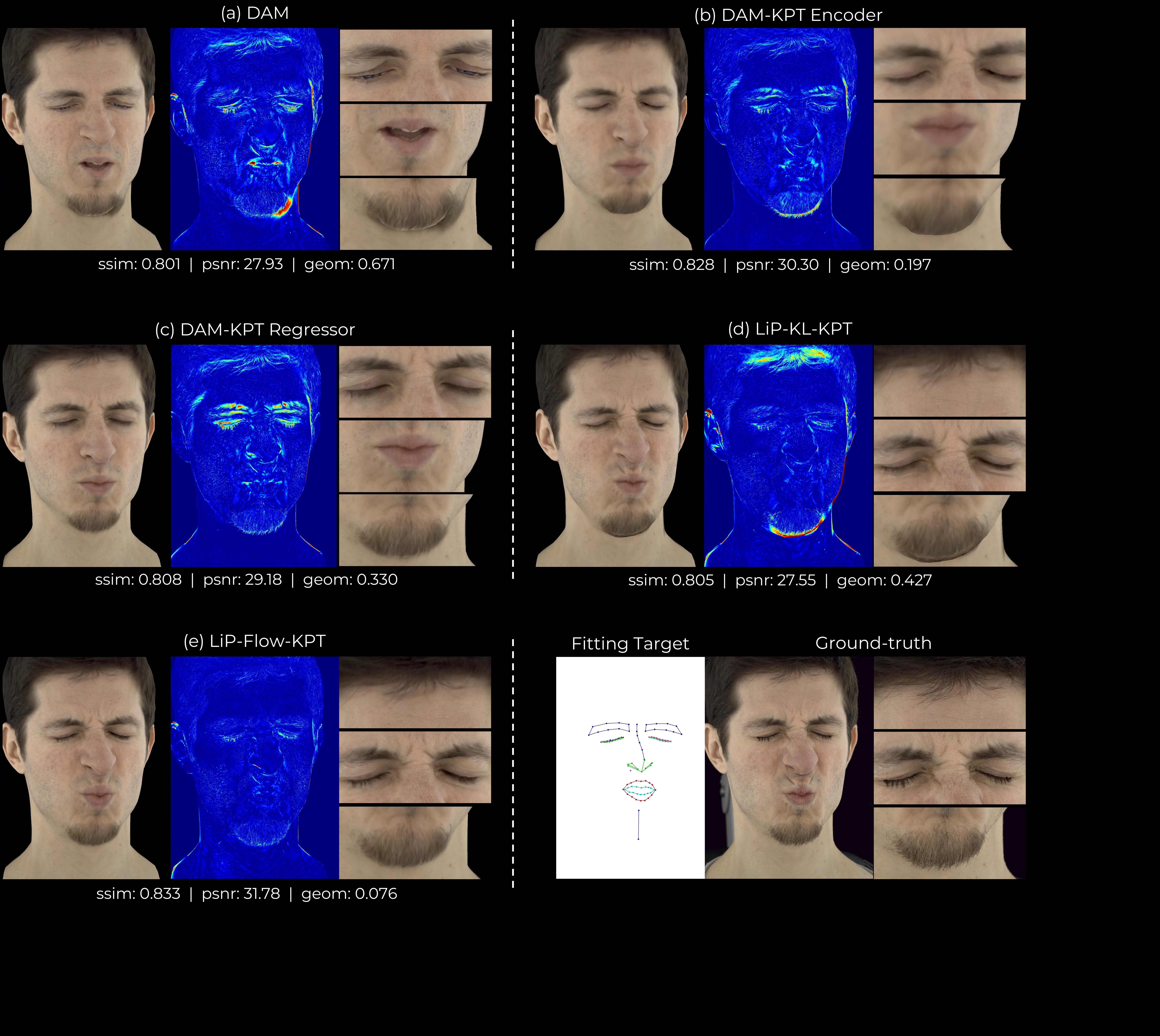}
\end{center}
\caption{\textbf{Reconstruction from 2D keypoints}. For all the models, we run inference-time optimization with sparse facial landmarks as the fitting target (cf. \Sec{kpt_fitting}). Note that full face is reconstructed from only the 2D keypoints. We visualize (left) frontal-view renderings of the optimized latent codes, (center) the difference between rendered and the ground-truth images and (right) close looks for three regions. Compared to the baselines, our model \modelname (e) achieves lower geometry error and reconstructs better textures, particularly with no errors around the chin and beard.  
}
\label{fig:qual_kpt_s1}
\end{figure*}

\begin{figure*}[t]
\begin{center}
\includegraphics[width=\linewidth]{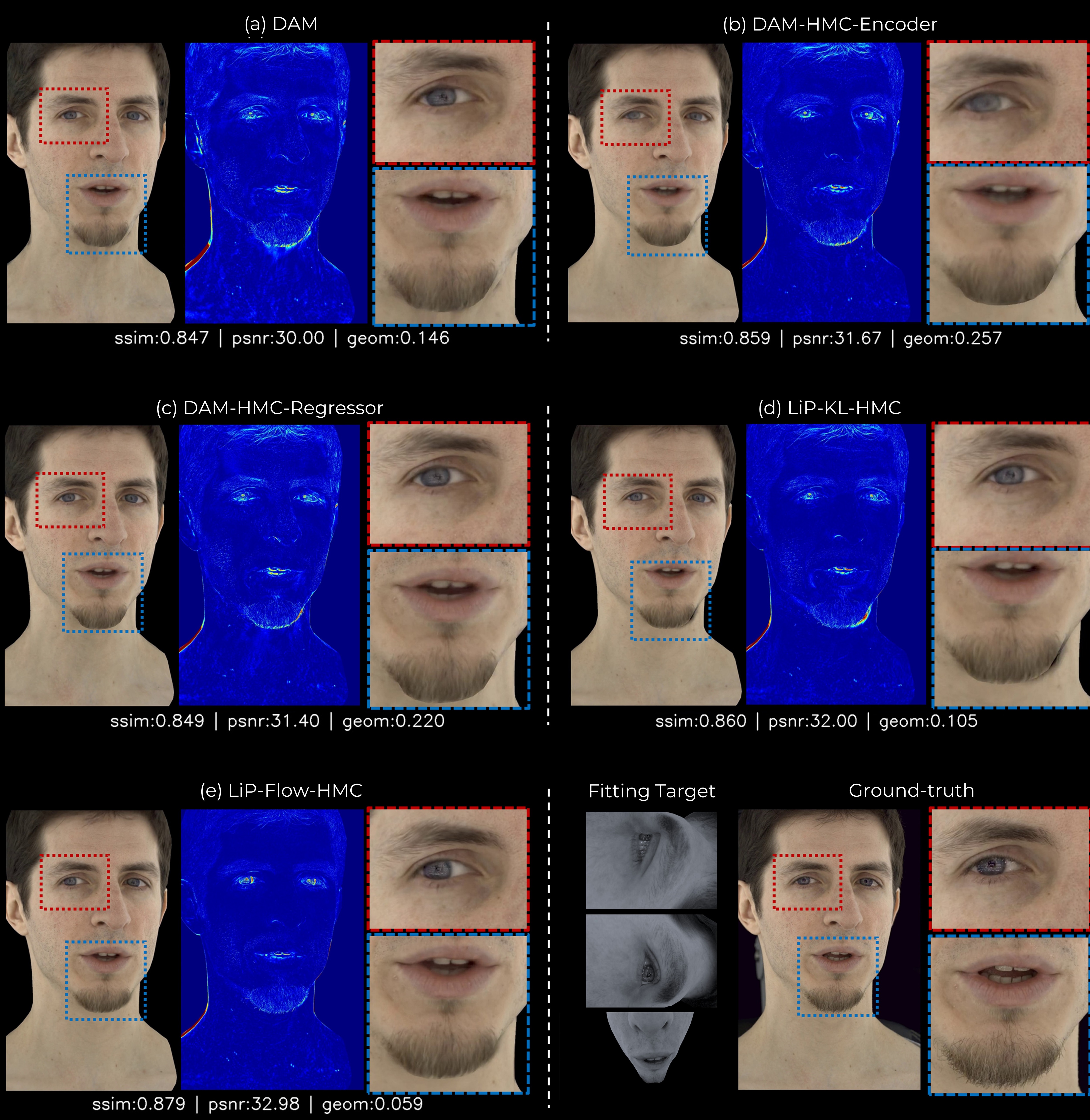}
\end{center}
\caption{\textbf{Reconstruction from HMC views}. For all the models, we run inference-time optimization with partial HMC views as the fitting target (cf. \Sec{hmc_fitting}). Note that full face is reconstructed from only the HCM images.
We visualize (left) frontal-view renderings of the optimized latent codes, (center) the difference between rendered and the ground-truth images and (right) close looks for two regions. Our model \modelname (e) achieves lower geometry error and reconstructs a sharper texture compared to the baselines.
}
\label{fig:qual_res_1}
\end{figure*}

\vspace{-0.2cm}
In Figures \ref{fig:qual_res_1} and \ref{fig:qual_kpt_s1}, we present qualitative results for both the HMC-view and 2D keypoint observations. For all the models, we optimize the latent codes via iterative fitting and present the frontal-view decodings. We provide animated flip-comparisons in our supplementary video.

\clearpage

\clearpage

\begin{figure}[t]
\begin{center}
\includegraphics[width=0.7\columnwidth]{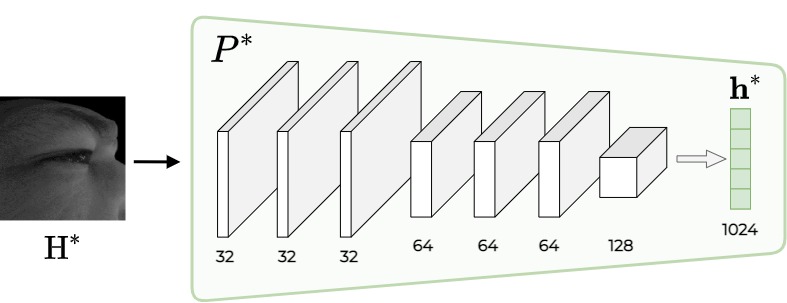}
\end{center}
\vspace{-0.3cm}
\caption{\textbf{HMC-encoder block for an input HMC image}. Number of output channels is provided in the figure.
}
\label{fig:sup_hmc_encoder}
\vspace{0.4cm}
\end{figure}

\section{Architecture Details}
\vspace{-0.2cm}
In our work, we choose the Deep Appearance Model (DAM) of Lombardi \etal \cite{lombardi2018deep} as the base model where the DAM-encoder (Q) and the decoder (D) networks use the same architecture and hyper-parameters. We provide details for the components introduced by us, namely the HMC-encoder ($P$) and the normalizing flow model in the latent space ($F$). We use PyTorch \cite{pytorch} for training our models. 

In terms of computational complexity, our latent flow requires $0.011$ GFLOPs to transform a latent sample. It is $1.736$ and $0.004$ for the HMC-Encoder and the KPT-Encoder, respectively. For reference, the DAM-Encoder and the decoder need $2.54$ and $2.0$ GFLOPs.
The computational requirement for our flow network and the KPT-Encoder are lower due to the lower dimensional inputs. We use the \texttt{fvcore}\footnote{\scriptsize\url{https://github.com/facebookresearch/fvcore/blob/main/docs/flop_count.md}} library for computing FLOPs.

\subsection{HMC-encoder}
Our HMC-encoder consists of $3$ separate convolutional blocks for each of the HMC image \small$\vecb{H}^* \in \real^{480 \times 640}$\normalsize. \Fig{sup_hmc_encoder} illustrates a convolutional block. We use convolutions with kernel size 4 and stride 2, followed by leaky relu activation functions \cite{maas2013rectifier} with a negative slope of $0.2$. The final output is reshaped into a $1024$-dimensional hidden vector $\vecb{h}^*$. 

After getting initial representations $\vecb{h}^*$ for $3$ HMC views, we apply a $3$-channel attention operation to calculate the HMC representation vector $\vecb{h} \in \real^{1024}$ by following \Eq{hmc_encoder_h}. We experimented with different alternatives such as concatenation of all the representations $\vecb{h}^*$ and experimentally verified that the attention mechanism yields the best performance. The attention weights $\vecb{W}^A \in \real^{1024 \times 3}$ are initialized with samples from $\mathcal{N}(0, 1)$. We finally estimate the mean $\bmP{\mu} \in \real^{256}$ and the standard deviation $\bmP{\sigma} \in \real^{256}$ by using linear layers without any activation.

\begin{figure}[t]
\begin{center}
\includegraphics[width=0.8\columnwidth]{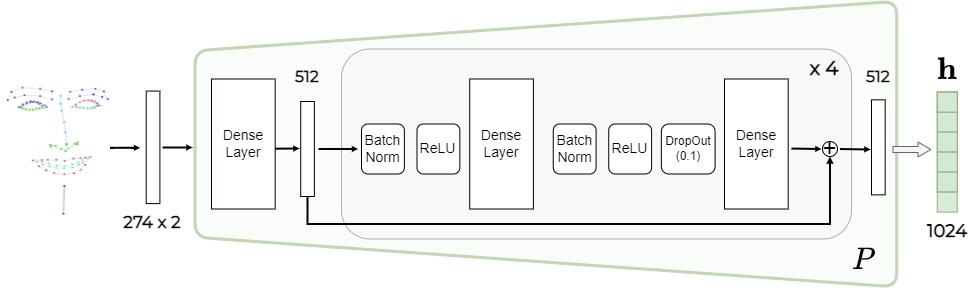}
\end{center}
\vspace{-0.3cm}
\caption{\textbf{KPT-encoder.} Input keypoints are of shape $274\times2$. We stack $4$ residual blocks with inputs and outputs of shape $512$. The final representation $\vecb{h}$ is a $1024$ dimensional vector.
}
\label{fig:sup_kpt_encoder}
\vspace{0.3cm}
\end{figure}

\begin{figure}[t]
  \begin{minipage}[c]{0.54\textwidth}
    \includegraphics[width=\textwidth]{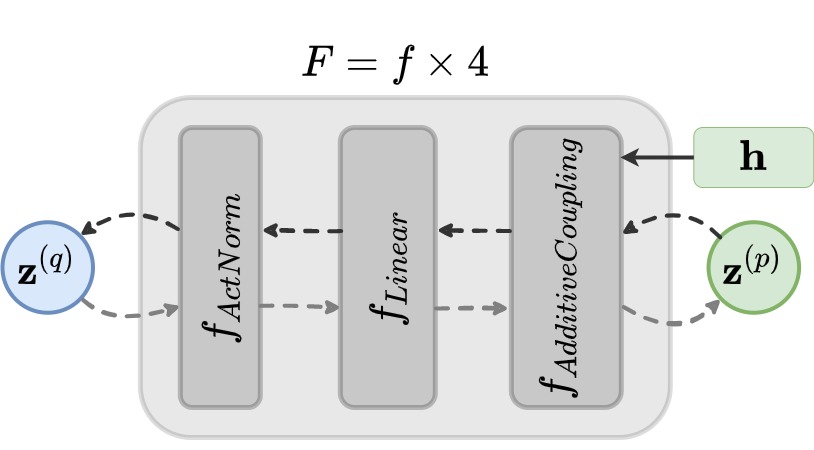}
  \end{minipage}\hfill
  \begin{minipage}{0.44\textwidth}
    \caption{\textbf{Overview of our normalizing flow network $F$}. We stack $4$ flow steps following \cite{kingma2018glow}. Each step consists of activation normalization, linear and additive coupling layers. The condition input $\vecb{h}$ is passed to the additive coupling layer at every level. $\vecb{h}$ corresponds to the HMC-view representations (cf. \Eq{hmc_encoder_h}).
    }
    \label{fig:sup_flow_architecture}
  \end{minipage}
\vspace{0.2cm}
\end{figure}

\subsection{KPT-encoder}
\label{sec:kpt_encoder_details}
Our KPT-encoder takes $274\times2$ dimensional keypoint inputs, which is mapped to an initial hidden representation of size $512$ by a dense layer. We then stack $4$ residual blocs consisting of batch normalization, ReLU, dense layers and a dropout (\Fig{sup_kpt_encoder}). Finally, the $512$ dimensional hidden representation is transformed to the keypoint representation vector $\vecb{h} \in \real^{1024}$ by a dense layer. 

\subsection{Latent Flow}
\vspace{-0.2cm}
We use the normalizing flow architecture proposed in \cite{kolotouros2021probabilistic}. Our flow network adapts building blocks of Glow \cite{kingma2018glow} to $1$D input where the invertible $1\times1$ convolution layer is replaced by a linear layer. We implement the normalizing flow network in our \modelname by using the \texttt{nflows} library \cite{nflows}. More specifically, we use \texttt{ActNorm}, \texttt{LULinear} and \texttt{AdditiveCouplingTransform} classes to implement the $f_{ActNorm}$, $f_{Linear}$ and $f_{AdditiveCoupling}$ layers in \Fig{sup_flow_architecture}. Hence, in \Tab{sup_flow_hyperparams}, we provide the hyper-parameters in the same interface with the \texttt{nflows} library. Our choice of the internal network \texttt{NN} for the coupling layer is a residual block conditioned on $\vecb{h}$ (\ie, context) as in \cite{kolotouros2021probabilistic}.

In our experiments, we find that the volume preserving additive coupling layer \cite{dinh2014nice} is essential. Our model \modelname often does not converge with the affine coupling layer. 

\section{Dataset}
We follow the same data preprocessing steps as in \cite{lombardi2018deep}. Each subject's data consists of $\sim$10,000 training frames where we have 40 camera views for each frame. We evaluate models on a held-out set of $\sim$500 frames. 

Our synthetic HMC dataset simulates the application conditions by applying augmentations including different lighting, background and headset orientation. The synthetic HMC images are generated by re-projecting the multi-view camera images into virtual head-mounted camera views.  More specifically, we sample headset camera parameters and render the corresponding HMC view by using the tracked mesh and the texture available in the multi-view dataset. We note that the synthetic and real HMC images are not photometrically alike due to the differences in the lighting conditions and camera properties. 

\begin{table}[t]
    \begin{minipage}[c]{0.55\textwidth}
	\center
	\renewcommand{\arraystretch}{1.1}
	\setlength\tabcolsep{7pt}%
	\begin{tabular}{l r}
		Hyper-parameter & Value \\
		\hline
		features & 256 \\
		hidden\_features & 1024 \\
		num\_layers & 4 \\
		num\_blocks\_per\_layer & 2 \\
		dropout\_probability & 0 \\
		activation & ReLU \\
		batch\_norm\_within\_layers & True\\
		\hline
	\end{tabular}
	\end{minipage}\hfill
    \begin{minipage}{0.45\textwidth}
	\caption{\textbf{Hyper-parameters of our normalizing flow network.}}
	\label{tab:sup_flow_hyperparams}
	\end{minipage}
	\vspace{0.5cm}
\end{table}

\section{Experiment Details}
\subsection{Training}
In our experiments, we use the same hyper-parameters for all the models as in \cite{lombardi2018deep}. The batch size and the learning rate are  $16$ and $5e^{-4}$, respectively. We allow models to train for $200,000$ steps. Different from \cite{lombardi2018deep}, we implement early stopping as we observe overfitting issues for all the models. Accordingly, if the training loss on the held-out set does not improve for $20,000$ steps, training is terminated. We use the following training objective (see \Eq{training_objective}):

\begin{align}
    \mathcal{L} &= \sum_{v} \lambda_{I}\mathcal{L}_{I} + \lambda_{M}\mathcal{L}_{M} + \lambda_{L}\mathcal{L}_{L},
    \label{eq:traininb_objective_supp}
\end{align}

In the training objective $\lambda_I = 10$, $\lambda_G = 1$. In the keypoint setting where we use the KPT-encoder conditioned on 2D keypoints, we observe very high amount of variance in the $P$ space. To alleviate this, we introduce an additional loss term to the training objective, penalizing the entropy of the predicted Gaussian $\gaussian(\bmP{\mu}, \bmP{\sigma})$. Its weight is the same as the latent log-likelihood term $\lambda_L$. Note that our LiP-Flow-KPT achieve the best results even in the absence of this prior entropy regularization term.

Since the latent regularization term $\mathcal{L}_L$ (\Eq{training_objective}) takes different forms in different models, its corresponding weight $\lambda_L$ is the most important hyper-parameter for the training. We run a mini hyper-parameter tuning on subject 1 and use the same set of values on the remaining subjects. 

\noindent\textbf{DAM} We considered KL-divergence (KL-D) weights of $0.01$ and $1$. We also followed an annealing strategy by starting it from $0.01$ and increasing it until $1$. In our experiments, the reported model achieved the best performance with the KL-D weight of $0.01$ as in \cite{lombardi2018deep}.
\\
\noindent\textbf{DAM-HMC-encoder} Unlike DAM, this model performed best when trained with the KL-D weight of $1$.
\\
\noindent\textbf{DAM-HMC-Regressor} In this setting, we use a pre-trained DAM-encoder to get the training labels and train the HMC-encoder with the log-likelihood objective. As it is the only training objective, we set the weight to $1$. 
\\
\noindent\textbf{LiP-KL} We followed the same hyper-parameter setup with the DAM. LiP-KL performs better when it is trained with KL-D weight $1$. 
\\
\noindent\textbf{LiP-Flow} Since we replace the KL-D term with a latent likelihood, we use a different set of weights. We consider weights ${0.005, 0.0001, 0.0005}$. Our model achieved the best performance when $\lambda_L = 0.0005$.

\subsection{Inference-time Optimization (\Sec{fitting_overview})}
We use the HMC-encoder to estimate a prior distribution which is then used for initialization and likelihood evaluation of the latent code. The HMC-encoder is available for all the models except the DAM (see \Fig{overview}). The prior distribution is set to Normal $\gaussianpri$ for the DAM baseline. 

In our evaluations, we assume that the camera parameters and the view-vector are available. The latent code is decoded into $3$D geometry and view-dependent texture which are then rendered to either frontal view or HMC views. Note that we decode the latent code $3$ times for each HMC-view if the target observations are the HMC images. We sample HMC camera parameters and the corresponding view-vector from an HMC dataset. 

The optimization objective (\Eq{fitting_objective}) consists of image reconstruction and latent likelihood terms. For the masked frontal-view and partial HMC-view targets, the image loss is calculated on the visible regions only. After setting the image loss weight to $1$, a grid search is performed on the first evaluation batch for the latent likelihood weight $\lambda_L$ and the learning rate. We then report the performance with the best performing hyper-parameters. The learning rate is determined from $\{0.01, 0.1, 0.5\}$. We use different sets for the latent likelihood weight $\lambda_L$ for the frontal-view and HMC-view targets, which are $\{0.01, 0.001, 0.0001\}$ and $\{0.1, 0.01, 0.001\}$, respectively. Since the fitting task is more challenging with the HMC-view targets, we observe that the models rely on the prior more and hence we run the hyper-parameter tuning with larger weights.

We use the ADAM algorithm \cite{kingma2014adam} for optimizing the latent code. The pretrained decoder is frozen during the optimization. We run the optimization up to $200$ steps and stop if there is no improvement in terms of the fitting objective. The average number of steps was $<200$ for all models. 

\section{Broader Impact Statement}
\label{sec:broaded_impact}
Since we aim to build photorealistic personalized avatars, our work constitutes a potential risk for negative use cases such as fake content generation and identity theft. Although the current state of the art including our work has not yet achieved photorealism, future research may achieve metric quality and produce synthetic data that are indistinguishable from the real ones. While synthesis of fake content is a problem for all generative models, our work as well as the prior works on 3D avatars present another potential misuse. By driving an established personalized avatar model, third parties may fake the identity. This can be prevented by introducing a verification layer to the mobile telepresence pipeline via retina-based bio-metric authentication systems. 

\end{document}